\documentclass[journal]{IEEEtran}
\ifCLASSINFOpdf
\else
\fi
\usepackage{amssymb}
\usepackage{slashbox}
\usepackage{mathrsfs}
\usepackage{mathrsfs}
\usepackage{mathrsfs}
\usepackage{amsfonts}
\usepackage{amsfonts}
\usepackage{amsfonts}
\usepackage{mathrsfs}
\usepackage{amsmath}
\usepackage{tabularx}

\usepackage{graphicx}
\usepackage{multirow}
\usepackage{caption2}
\usepackage{float}
\usepackage{booktabs}
\usepackage{algorithm}
\usepackage{algorithmic}
\usepackage{subfigure}

\newtheorem{theorem}{Theorem}[section]
\renewcommand{\theequation}{\thesection.\arabic{equation}}
\numberwithin{equation}{section}

\newtheorem{assumption}[theorem]{Assumption}

\newtheorem{definition}[theorem]{Definition}

\newtheorem{lemma}[theorem]{Lemma}

\newtheorem{proposition}[theorem]{Proposition}
\newtheorem{remark}[theorem]{Remark}

\begin{document}
%
\title{Shrinkage  degree  in $L_2$-re-scale  boosting for regression}
%
%
%

\author{Lin Xu,~Shaobo Lin,~Yao Wang
        and~Zongben Xu
\IEEEcompsocitemizethanks{\IEEEcompsocthanksitem L. Xu, Y. Wang,
and Z. Xu are  with the Institute for Information and System
Sciences, School of Mathematics and Statistics, Xi'an Jiaotong
University, Xi'an 710049, China; S. Lin is with the College of Mathematics and Information Science, Wenzhou
University, Wenzhou 325035, China}}

\maketitle

\begin{abstract}
Re-scale boosting (RBoosting) is a variant of  boosting which can
essentially improve the generalization performance of boosting learning. The
key feature of RBoosting lies in introducing a shrinkage degree to
re-scale the ensemble estimate in each gradient-descent step. Thus,
the shrinkage degree determines the performance of RBoosting.
 The aim of this paper is to develop a concrete analysis concerning
how to determine the shrinkage degree in $L_2$-RBoosting. We propose
two feasible ways to select the shrinkage degree. The first one is to
parameterize  the shrinkage degree and the other one is to develope a
data-driven approach of it. After rigorously analyzing the
importance of the shrinkage degree in  $L_2$-RBoosting learning, we
compare the pros and cons of the proposed methods. We find that
although these approaches can  reach the same learning rates, the
structure of the final estimate of the parameterized approach is
better, which sometimes yields a better generalization capability
when the number of sample is finite. With this, we  recommend to
parameterize the shrinkage degree of $L_2$-RBoosting. To this end,
we present an adaptive parameter-selection strategy for shrinkage
degree and verify its feasibility through both theoretical analysis  and
numerical verification.
  The obtained results enhance the understanding of RBoosting and further give
guidance on how to use $L_2$-RBoosting for regression tasks.
\end{abstract}


%
\IEEEpeerreviewmaketitle

\section{Introduction}
%
%
%
%
\IEEEPARstart{B}{oosting} is a learning system which combines
many parsimonious models to produce a  model with  prominent
predictive performance. The underlying intuition is that combines
many rough rules of thumb can  yield a good composite learner. From
the statistical viewpoint,  boosting   can be viewed as a form of
functional gradient decent \cite{Friendman2001}. It connects various
boosting algorithms to optimization problems with specific loss
functions. Typically, $L_2$-Boosting \cite{Buhlmann2003,Freund1996}
can be interpreted as an stepwise additive learning scheme that
concerns the problem of minimizing the $L_2$ risk. Boosting is
resistant to overfitting \cite{Friendman2000} and thus, has
triggered enormous research activities in the past twenty years
\cite{Buhlmann2007,Duffy2002,Freund1995,Friendman2001,Schapire1990}.

Although the universal consistency of boosting has already been
verified in \cite{Bartlett2007},   the numerical  convergence rate
of boosting is a bit slow \cite{Bartlett2007,Livshits2009}. The main
reason for such a drawback is that
  the step-size derived via linear search in boosting can not always guarantee
the most appropriate one \cite{Efron2004,Lin2015}. Under this
circumstance,   various variants of boosting, comprising the
regularized boosting via shrinkage (RSBoosting)
\cite{Ehrlinger2012}, regularized boosting via truncation
(RTBoosting) \cite{Zhang2005} and $\varepsilon$-Boosting
\cite{Hastie2007} have been developed via introducing additional
parameters to control the step-size. Both experimental and
theoretical results
\cite{Friendman2001,Ehrlinger2012,Zhao2007,Buhlmann2007} showed that
these variants outperform the classical boosting within a certain
extent. However, it also needs verifying whether the learning
performances of these variants can be further improved, say, to the
best of our knowledge, there is not any related theoretical analysis
 to illustrate the optimality of these variants, at least for a certain aspect, such as the
generalization capability, population (or numerical) convergence rate, etc.

Motivated by the recent development of relaxed greedy algorithm \cite{Temlyakov2008} and
sequential greedy algorithm \cite{Zhang2003}, Lin et al.
\cite{Lin2015} introduced a new variant of boosting named as the
re-scale boosting (RBoosting). Different from the existing variants
that focus on controlling the step-size, RBoosting builds upon
re-scaling the ensemble  estimate and implementing the linear search
without any restrictions on the step-size in each gradient descent
step. Under such a setting, the optimality of the population
convergence rate of RBoosting was verified. Consequently, a tighter
generalization error of RBoosting was deduced. Both theoretical
analysis and experimental results in \cite{Lin2015} implied that
RBoosting is better than  boosting, at least for the $L_2$ loss.

As there is no free lunch, all the variants improve the learning
performance of boosting at the cost of   introducing an additional
parameter, such as the truncated parameter in RTBoosting,
regularization parameter in RSBoosting,  $\varepsilon$ in
$\varepsilon$-Boosting, and shrinkage degree in RBoosting. To
facilitate the use of these variants, one should also present
strategies to   select such parameters. In particular,  Elith et al.
\cite{Elith2008} showed that $ 0.1$ is a feasible choice of
$\varepsilon$ in $\varepsilon$-Boosting; B\"{u}hlmann and Hothorn
\cite{Buhlmann2007} recommended the selection of $0.1$ for the
regularization parameter in RSBoosting; Zhang and Yu
\cite{Zhang2005} proved that $\mathcal O(k^{-2/3})$ is a good value
of the truncated parameter in RTBoosting, where $k$ is the number of
iterations. Thus, it is interesting  and important to provide a
feasible  strategy for selecting  shrinkage degree in RBoosting.

%
%
%
%
%
%
%

Our aim in the current article is to propose several feasible
strategies to select the shrinkage degree in $L_2$-RBoosting and
analyze their pros and cons. For this purpose, we need to justify
the essential role  of the shrinkage degree in $L_2$-RBoosting.
After rigorously theoretical analysis, we find that, different from
other parameters such as the truncated value, regularization
parameter, and $\varepsilon$ value, the shrinkage degree does not
affect the learning rate, in the sense that, for arbitrary finite
shrinkage degree, the learning rate of corresponding $L_2$-RBoosting
can reach the existing best record of all boosting type algorithms.
This means that if the number of samples is infinite, the shrinkage
degree does not affect the generalization capability of
$L_2$-RBoosting. However, our result also shows that the essential
role of the shrinkage degree in $L_2$-RBoosting lies in its
important impact on the constant of the generalization error, which
is crucial when there are only finite number of  samples.
 In such a sense, we  theoretically proved that there exists an optimal
shrinkage degree to minimize the generalization error of
$L_2$-RBoosting.

We then aim to develop two effective methods for a ``right'' value
of the shrinkage degree. The first one is to consider the shrinkage
degree as a parameter in the learning process of $L_2$-RBoosting.
The other one is to learn the shrinkage degree from the  samples
directly and we call it as the  $L_2$ data-driven RBoosting
($L_2$-DDRBoosting). We find that the above two approaches can reach
the same learning rate and the number of parameters in
$L_2$-DDRBoosting is less than that of  $L_2$-RBoosting. However, we
also prove that the estimate deduced from $L_2$-RBoosting possesses
a better structure (smaller $l^1$ norm), which sometimes leads a
much better generalization capability for some special weak
learners. Thus, we recommend the use of $L_2$-RBoosting in practice.
Finally,  we develop an adaptive shrinkage degree selection strategy
for $L_2$-RBoosting. Both the theoretical and experimental results
verify the feasibility and outperformance of $L_2$-RBoosting.

The rest of paper is organized as  follows. In Section 2, we
give a brief introduction to  the $L_2$-Boosting, $L_2$-RBoosting and $L_2$-DDRBoosting.
In Section 3, we study the related theoretical behaviors of
$L_2$-RBoosting. In Section 4, a series of simulations and
real data experiments are employed to illustrate our theoretical assertions. In
Section 5, we provide the proof of the main results. In the last
section, we draw a simple conclusion.

\section{$L_2$-Boosting, $L_2$-RBoosting and $L_2$-DDRBoosting}

Ensemble techniques such as bagging \cite{Breiman1996}, boosting
\cite{Freund1995}, stacking \cite{Smyth1999}, Bayesian  averaging
\cite{Mackay1991} and random forest \cite{Breiman2001} can
significantly improve performance in practice and benefit from
favorable learning capability. In particular, boosting and its
variants are based on a rich theoretical analysis, to just name a
few,  \cite{Bagirov2010}, \cite{Bartlett2007}, \cite{Bickel2006},
\cite{Buhlmann2003}, \cite{Friendman2000}, \cite{Lin2013},
\cite{Lin2015}, \cite{Zhang2005}. The aim of this section is to
introduce some concrete boosting-type learning schemes for
regression.

In a regression problem  with  a covariate  $X$ on $\mathcal
X\subseteq\mathbf R^d$ and a real response variable $Y\in\mathcal
Y\subseteq \mathbf R$, we observe $m$ i.i.d. samples
$D_m=\{(x_i,y_i)\}_{i=1}^m$ from an unknown underlying distribution
$\rho$. Without loss of generality, we always assume $\mathcal
Y\subseteq [-M,M]$, where $M<\infty$ is a positive real number. The
aim is to find a function to minimize the generalization error
$$
          \mathcal E(f)=\int\phi(f(x),y)d\rho,
$$
where $\phi:\mathbf R\times\mathbf R\rightarrow\mathbf R_+$ is
called a loss function \cite{Zhang2005}. If
$\phi(f(x),y)=(f(x)-y)^2$, then the known regression function
$$
        f_\rho(x)=\mathbf E\{Y|X=x\}
$$
minimizes the generalization error. In such a setting, one is
interested in finding a function $f_D$ based on $D_m$ such that
$\mathcal E(f_D)-\mathcal E(f_\rho)$ is small. Previous study
\cite{Buhlmann2003} showed that $L_2$-Boosting can successfully
tackle this problem.

  Let $
          S=\{g_1,\dots,g_n\}
$
 be the set of weak learners (regressors)  and define
$$
          \mbox{span}(S)=\left\{\sum_{j=1}^na_jg_j:g_j\in S,
          a_j\in\mathbf R, n\in\mathbf N\right\}.
$$
Let
$$
           \|f\|_m=\sqrt{\frac1m\sum_{i=1}^mf(x_i)^2}, \ \mbox{and}\
           \langle f,g\rangle _m=\frac1m\sum_{i=1}^mf(x_i)g(x_i)
$$
be the empirical norm and empirical inner product, respectively.
Furthermore, we define the empirical risk as
$$
      \mathcal E_D(f)=\frac1m\sum_{i=1}^m|f(x_i)-y_i|^2.
$$
Then the gradient descent view of $L_2$-Boosting
\cite{Friendman2001} can be interpreted as follows.

\begin{algorithm}[H]\caption{Boosting}\label{alg1}
\begin{algorithmic}
\STATE {{ Step 1(Initialization)}: Given data
$\{(x_i,y_i):i=1,\dots,m\}$, dictionary $S$, iteration number $k^*$
and $f_0\in\mbox{span}(S)$}. \STATE{ { Step 2(Projection of gradient
)}: Find $g_k^*\in S$ such that
$$
            g_k^* = \arg {\max _{g \in S}}|{\langle {r_{k - 1}},g\rangle _m}|,
$$
where residual $r_{k-1}=y-f_{k-1}$ and $y$ is a function satisfying
$y(x_i)=y_i$. }

 \STATE{{Step 3(Linear search)}:
$$
            f_{k}=f_{k-1}+\langle r_{k-1},g_k^*\rangle_mg_k^*.
$$
}
 \STATE{ { Step 4
(Iteration)} Increase $k$ by one and repeat Step 2 and Step 3 if
$k<k^*$.}
\end{algorithmic}
\end{algorithm}

\begin{remark}
In the step 3 in Algorithm \ref{alg1}, it is easy to check that
$$
            \langle r_{k-1},g_k^*\rangle_m=\arg\min_{\beta_k \in
            \mathbf R}\mathcal E_D(f_{k-1}+\beta_k g_k^*).
$$
Therefore, we call it as the linear search step.
\end{remark}

In spite of $L_2$-Boosting  was proved to be consistent
\cite{Bartlett2007} and overfitting resistance \cite{Buhlmann2003},
multiple studies \cite{DeVore1996,Livshits2009,Temlyakov2008a}
also showed that its population convergence rate is far slower than
the best nonlinear approximant. The main reason is that the linear
search in Algorithm \ref{alg1} makes $f_{k+1}$ to be not always the greediest
one \cite{Efron2004,Lin2015}. Hence, an advisable method is to
control the step-size in the linear search step of Algorithm \ref{alg1}.
Thus, various variants of boosting, such as the
$\varepsilon$-Boosting \cite{Hastie2007} which specifies the
step-size as a fixed small positive number $\varepsilon$ rather than
using the linear search, RSBoosting\cite{Ehrlinger2012} which
multiplies a small regularized factor to the step-size deduced from
the linear search and RTBoosting\cite{Zhang2005} which truncates the
linear search in a small interval have been developed. It is obvious
that the core difficulty of these schemes roots in how to select an
appropriate step-size. If the step size is too large, then these
algorithms may face the same problem as that of Algorithm \ref{alg1}. If
the step size is too small, then the population convergence rate is
also fairly slow.

Other than the aforementioned strategies that focus on controlling
the step-size of  $g_k^*$, Lin et al. \cite{Lin2015} also derived a
new backward type strategy, called the re-scale boosting
(RBoosting), to improve the population convergence rate and
consequently, the generalization capability of boosting. The core
idea is that if the approximation (or learning) effect of the $k$-th
iteration may not work as expected, then  $f_k$ is regarded  to be
too aggressive. That is, if a new iteration is employed, then   the
previous estimator $f_k$ should be re-scaled. The following
Algorithm \ref{alg2} depicts the main idea of $L_2$-RBoosting.

\begin{algorithm}[H]\caption{RBoosting}\label{alg2}
\begin{algorithmic}
\STATE {{ Step 1(Initialization)}: Given data
$\{(x_i,y_i):i=1,\dots,m\}$, dictionary $S$, a set of shrinkage
degree $\{\alpha_k\}^{k^*}_{k=1}$ where $\alpha_k=2/(k+u), u \in \mathbf{N}$,  iteration number $k^*$
and $f_0\in\mbox{span}(S)$}.
 \STATE{ { Step 2(Projection of gradient)}: Find $g_k^*\in S$ such that
$$
            g_k^* = \arg {\max _{g \in S}}|{\langle {r_{k - 1}},g\rangle _m}|,
$$ where the residual $r_{k-1}=y-f_{k-1}$ and $y$ is a function satisfying
$y(x_i)=y_i$.
  }
 \STATE{{Step 3( Re-scaled linear search)}:
 $$
            f_{k}=(1-\alpha_k)f_{k-1}+\langle
            r^s_{k-1},g_k^*\rangle_mg_k^*,
$$
where the shrinkage residual
 $r^s_{k-1}=y-(1-\alpha_k)f_{k-1}$. }
 \STATE{ { Step 4
(Iteration)}: Increase $k$ by one and repeat Step 2 and Step 3 if
$k< k^*$.}
\end{algorithmic}
\end{algorithm}

\begin{remark}
It is easy to see that
$$
            \langle r^s_{k-1},g_k^*\rangle_m=\arg\min_{\beta_k\in
            \mathbf R}\mathcal E_D((1-\alpha_k)f_{k-1}+\beta_k g_k^*).
$$
This is the only difference between boosting and RBoosting. Here we
call $\alpha_k$ as the shrinkage degree. It can be found  in the
above Algorithm \ref{alg2} that the shrinkage degree is considered  as a
parameter.
\end{remark}

$L_2$-RBoosting  stems from  the ``greedy algorithm with fixed
relaxation'' \cite{Temlyakov2008a} in nonlinear approximation. It is
different from the $L_2$-Boosting algorithm proposed in
\cite{Bagirov2010}, which adopts the idea of ``$X$-greedy algorithm
with relaxation''  \cite{Barron2008}. In particular, we employ
$r_{k-1}$ in Step 2 to represent residual rather than the shrinkage
residual $r^s_{k-1}$ in Step 3. Such a difference makes the design
principles of RBoosting and the boosting algorithm in
\cite{Bagirov2010} to be totally distinct. In RBoosting, the
algorithm comprises two steps: the projection of gradient step to
find the optimum weak learner $g_k^*$ and the re-scale linear search
step to fix its step-size $\beta_k$. However, the boosting algorithm
in \cite{Bagirov2010} only concerns the optimization problem
$$
         \arg\min_{g^*_k\in S,\beta_k \in\mathbf
         R}\|(1-\alpha_k)f_{k-1}+\beta_k g^*_k\|_m^2.
$$
The main drawback is, to the best of our knowledge, the closed-form
solution of the above optimization problem only holds for the $L_2$
loss. When faced with other loss, the boosting algorithm in
\cite{Bagirov2010} cannot be efficiently numerical solved. However,
it can be found in \cite{Lin2015} that RBoosting is feasible for
arbitrary loss. We are currently studying the more concrete
comparison study between these two re-scale boosting algorithms
\cite{Xu2015}.

It is known that $L_2$-RBoosting  can improve the  population convergence rate and
generalization capability  of   $L_2$-Boosting \cite{Lin2015}, but
the price is that there is an additional parameter, the shrinkage
degree $\alpha_k$,   just like the step-size parameter $\varepsilon$
in $\varepsilon$-Boosting \cite{Hastie2007},   regularized parameter
$v$ in  RSBoosting \cite{Ehrlinger2012} and truncated parameter $T$
in RTBoosting \cite{Zhang2005}. Therefore, it is urgent   to develop
a feasible  method to select the shrinkage degree. There are two
ways to choose a good shrinkage degree value. The first  one  is to
parameterize the shrinkage degree as in
Algorithm \ref{alg2}. We set the shrinkage degree $\alpha_k=2/(k+u)$ and hope
to choose an appropriate value of $u$ via a certain
parameter-selection strategy. The other one  is to  learn the
shrinkage degree $\alpha_k$ from the samples directly. As we are
only concerned  with $L_2$-RBoosting in present paper, this idea can
be primitively realized by the following Algorithm \ref{alg3}, which is called as the data-driven RBoosting (DDRBoosting).

\begin{algorithm}[H]\caption{DDRBoosting}\label{alg3}
\begin{algorithmic}
\STATE {{ Step 1(Initialization)}: Given data
$\{(x_i,y_i):i=1,\dots,m\}$, dictionary $S$,iteration number $k^*$
and $f'_0\in\mbox{span}(S)$}.
 \STATE{ { Step 2(Projection of gradient)}: Find $g_k^*\in S$ such that
$$
            g_k^* = \arg {\max _{g \in S}}|{\langle {r_{k - 1}},g\rangle _m}|,
$$ where residual $r_{k-1}=y-f'_{k-1}$ and $y$ is a function satisfying
$y(x_i)=y_i$.}
 \STATE{{Step 3(Two dimensional linear search)}:
 Find $\alpha_k^*$ and $\beta_k^*\in\mathbf R$  such that
$$
      \mathcal E_D((1-\alpha_k^*)f'_{k-1}+\beta_{k}^*g^*_{k})=
      \inf_{(\alpha_k,\beta_k)\in\mathbf R^2}
      \mathcal E_D((1-\alpha_k) f'_{k-1}+\beta_kg^*_k)
$$
Update $f'_{k}=(1-\alpha_k^*)f'_{k-1}+\beta_k^*g_k^*$.}
 \STATE{ { Step 4
(Iteration)}: Increase $k$ by one and repeat Step 2 and Step 3 if
$k< k^*$.}
\end{algorithmic}
\end{algorithm}

The above Algorithm \ref{alg3} is motivated by the ``greedy algorithm with
free relaxation'' \cite{Temlyakov2012}. As far as  the $L_2$ loss is
concerned, it is easy to deduce the close-form representation of
$f'_{k+1}$ \cite{Temlyakov2008a}. However, for other loss functions,
we have not found any papers concerning the solvability of the
optimization problem in step 3 of the Algorithm \ref{alg3}.



\section{Theoretical behaviors}
In this section, we present  some theoretical results concerning the shrinkage degree. Firstly, we  study the relationship between shrinkage degree
and  generalization capability  in $L_2$-RBoosting.
The theoretical results reveal that the shrinkage degree plays a
crucial role in $L_2$-RBoosting  for regression with finite samples.
Secondly, we analyze the pros and cons of
 $L_2$-RBoosting and $L_2$-DDRBoosting. It is shown that the potential performance of
$L_2$-RBoosting is somewhat better than that of $L_2$-DDRBoosting.
Finally,  we   propose an adaptive parameter-selection strategy for
the shrinkage degree and theoretically verify its feasibility.

\subsection{Relationship between the generalization capability and  shrinkage degree}

At first, we give a few notations and concepts, which will be used
throughout the paper.
Let $\mathcal L_1(S):=\{f:f=\sum_{g\in S}a_gg\}$ endowed with the
norm
$$
     \|f\|_{\mathcal L_1(S)}:=\inf\left\{\sum_{g\in
S}|a_g|:f=\sum_{g\in S}a_gg\right\}.
$$
For $r>0$, the space
$\mathcal L_1^r$ is defined to be the set of all functions $f$ such
that, there exists $h\in\mbox{span}\{S\}$ such that
\begin{equation}\label{prior}
           \|h\|_{\mathcal L_1(S)}\leq\mathcal B, \ \mbox{and}\
           \|f - h\| \leq {\mathcal  B}{n^{ - r}},
\end{equation}
where $\|\cdot\|$ denotes the uniform norm for the continuous
function space $C(\mathcal X)$. The infimum of all such $\mathcal B$
defines a norm for $f$ on $\mathcal L_1^r$. It follows from
\cite{Barron2008} that  (\ref{prior}) defines an interpolation space
which has been widely used in nonlinear approximation
\cite{Barron2008,Lin2013,Temlyakov2008a}.

Let $\pi_Mt$ denote  the clipped value of $t$ at $\pm M$, that is,
$\pi_Mt:=\min\{M,|t|\}\mbox{sgn}(t)$. Then it is obvious that
\cite{Zhou2006}   for all $t\in\mathbf R$ and $y\in[-M,M]$ there
holds
$$
{\cal E}({\pi _M}{f_k}) - {\cal E}({f_\rho }) \le {\cal E}({f_k}) - {\cal E}({f_\rho }).
$$

By the help of the above descriptions, we are now in a position to
present the following Theorem \ref{THEOREM1}, which depicts the role
that the shrinkage degree  plays in $L_2$-RBoosting.

\begin{theorem}\label{THEOREM1}
Let $0<t<1$,  and $f_{k}$  be the estimate defined in Algorithm 2.
If $f_\rho\in \mathcal L_1^r$,  then  for arbitrary $k,u\in N$,
\begin{eqnarray*}
\begin{aligned}
           & {\mathcal E}({\pi _M}f_{ k}) - {\mathcal E}({f_\rho }
            ) \leq \\
           & \!C\! (M\!+\!\mathcal B)\!^2\left(\!2^\frac{3u^2+14u+20}{8u+8}k^{-1}
             \! +  \!  (m/k)^{-1}\log m\log\frac2t+n^{-2r}\right)
\end{aligned}
\end{eqnarray*}
holds  with probability at least $1-t$, where $C$ is a positive
constant depending only on $d$.
\end{theorem}

Let us first give some remarks of Theorem \ref{THEOREM1}. If we set
the number of iterations and the size of dictionary to satisfy $ k=
\mathcal O(m^{1/2})$, and $n \geq \mathcal O({m^{\frac{1}{{4r}}}})$,
then we can deduce a learning rate of $ \pi_Mf_k$  asymptotically as
$\mathcal O(m^{-1/2}\log m)$. This rate is independent of the
dimension and is the same as the optimal ``record'' for greedy
learning \cite{Barron2008} and boosting-type algorithms
\cite{Zhang2005}. Furthermore, under the same assumptions, this rate
is faster than those of boosting \cite{Bartlett2007} and RTBoosting
\cite{Zhang2005}. Thus, we can draw a rough conclusion that the
learning rate deduced in Theorem \ref{THEOREM1} is tight. Under this
circumstance, we think it can reveal the essential performance of
$L_2$-RBoosting.

Then, it can be found in Theorem \ref{THEOREM1} that if $u$ is
finite and the number of samples is infinite, the shrinkage degree
$u$ does not affect the learning rate of $L_2$-RBoosting, which means
its generalization capability is independent of $u$. However, it is
known that in the real world application, there are only finite number of samples available. Thus, $u$ plays a crucial role in the
learning process of $L_2$-RBoosting in practice. Our results in
Theorem \ref{THEOREM1} implies two simple guidance to deepen
the understanding of $L_2$-RBoosting.  The first one is that there
indeed exist an optimal $u$ (may be not unique) minimizing the generalization
error of $L_2$-RBoosting. Specifically, we can deduce a concrete
value of optimal $u$ via minimizing $\frac{3u^2+14u+20}{8u+8}$. As
it is very difficult to  prove the optimality of the constant, we
think it is more reasonable to reveal a rough trend  for choosing $u$
rather than providing a concrete value.
 The other one is that when
$u\rightarrow\infty$, $L_2$-RBoosting behaves as $L_2$-Boosting, the learning
rate cannot achieve $\mathcal O(m^{-1/2}\log m)$. Thus, we indeed
present a theoretical verification that $L_2$-RBoosting outperforms
$L_2$-Boosting.

\subsection{Pros and cons of $L_2$-RBoosting and $L_2$-DDRBoosting }

There is only one parameter, $k^*$, in $L_2$-DDRBoosting, as showed
in Algorithm \ref{alg3}.  This implies that $L_2$-DDRBoosting improves the
performance of $L_2$-Boosting without tuning another additional
parameter $\alpha_k$, which is superior to the other  variants of
boosting. The following Theorem \ref{THEOREM2} further shows that, as
the same as $L_2$-RBoosting, $L_2$-DDRBoosting can also improve the
generalization capability of $L_2$-Boosting.

\begin{theorem}\label{THEOREM2}
Let $0<t<1$,  and $f'_{k}$  be the  estimate defined in Algorithm \ref{alg3}.
If $f_\rho\in \mathcal L_1^r$, then  for any arbitrary $k\in N$,
\begin{equation*}
\begin{aligned}
        &    {\mathcal E}({\pi _M}f'_{k}) - {\mathcal E}({f_\rho }
            ) \leq \\ & C (M+\mathcal B)^2  \left(k^{-1}
             +    (m/k)^{-1}\log m\log\frac2t +n^{-2r}\right)
\end{aligned}
\end{equation*}
holds  with probability at least $1-t$, where $C$ is a constant
depending only on $d$.
\end{theorem}

By Theorem \ref{THEOREM2}, it seems that $L_2$-DDRBoosting can
perfectly solve the parameter selection problem in the re-scale-type
boosting algorithm. However, we also show in the following that
compared with $L_2$-DDRBoosting, $L_2$-RBoosting possesses an
important advantage, which is crucial to guaranteeing  the
outperformance  of $L_2$-RBoosting. In fact, noting that
$L_2$-DDRBoosting depends on a two dimensional linear search problem
(step 3 in Algorithm \ref{alg3}), the structure of the estimate ($\mathcal
L_1$ norm), can not always be good. If the  estimate
$f^\prime_{k-1}$ and $g_k^*$ are almost linear dependent, then the
values of $\alpha_k$ and $\beta_k$ may be very large, which
automatically leads a huge $\mathcal L_1$ norm of $f'_k$. We show in
the following Proposition \ref{Proposition1} that $L_2$-RBoosting
can avoid this phenomenon.

\begin{proposition}\label{Proposition1}
  If the $f_{k}$ is the  estimate defined in Algorithm \ref{alg2}, then there
  holds
$$
        \|f_{k}\|_{\mathcal L_1(S)}\leq C((M+\|h\|_{\mathcal L_1(S)})k^{1/2}+kn^{-r}).
$$
\end{proposition}

Proposition \ref{Proposition1} implies the estimate defined in
Algorithm \ref{alg2} possesses a controllable structure. This may
significantly improve the learning performance of $L_2$-RBoosting when
faced with some specified weak learners. For this purpose, we need to
introduce some definitions and conditions to qualify the weak
learners.

\begin{definition}
 Let $(\mathcal M,d)$ be a pseudo-metric space
and $T\subset\mathcal M$ a subset. For every $\varepsilon>0$, the
covering number $\mathcal N(T,\varepsilon,d)$ of $T$ with respect to
$\varepsilon$ and $d$ is defined as the minimal number of balls of
radius $\varepsilon$ whose union covers $T$, that is,
$$
                 \mathcal N(T,\varepsilon,d):=\min\left\{l\in\mathbf
                 N: T\subset\bigcup_{j=1}^lB(t_j,\varepsilon)\right\}
$$
for some $\{t_j\}_{j=1}^l\subset\mathcal M$, where
$B(t_j,\varepsilon)=\{t\in\mathcal M:d(t,t_j)\leq\varepsilon\}$.
\end{definition}

The $l_2$-empirical covering number of a function set is defined by
means of the normalized $l_2$-metric $d_2$ on the Euclidean space
$\mathbf R^d$ given in \cite{Shi2011} with $
                  d_2({\bf
                  a,b})=\left(\frac1m\sum_{i=1}^m|a_i-b_i|^2\right)^\frac12
$
 for  ${\bf a}=(a_i)_{i=1}^m, {\bf
                  b}=(b_i)_{i=1}^m\in\mathbf R^m.$
\begin{definition}
 Let $\mathcal F$ be a set of functions on $X$,
${\bf x}=(x_i)_{i=1}^m\subset X^m$,  and let
\[
\mathcal F|_{\bf x}:=\{(f(x_i))_{i=1}^m:f\in\mathcal F\}\subset R^m.
\]
 Set $\mathcal
N_{2,{\bf x}}(\mathcal F,\varepsilon)=\mathcal N(\mathcal F|_{\bf
x},\varepsilon,d_2)$. The $l_2$-empirical covering number of
$\mathcal F$ is defined by
$$
                 \mathcal N_2(\mathcal
                 F,\varepsilon):=\sup_{m\in\mathbf N}\sup_{{\bf
                 x}\in S^m}\mathcal N_{2,{\bf x}}(\mathcal
                 F,\varepsilon),\ \ \varepsilon>0.
$$
\end{definition}

%

Before presenting the main result in the subsection, we shall
introduce the following Assumption \ref{Assumption1}.

\begin{assumption}\label{Assumption1}
Assume the $l_2$-empirical covering number of $\mbox{span} (S)$
satisfies
$$
                  \log\mathcal N_2(B_1,\varepsilon)\leq
                  \mathcal L\varepsilon^{-\mu},\ \ \forall\varepsilon>0,
$$
where
$$
            B_R=\{f\in \mbox{span}(S):\|f\|_{\mathcal L^1(S)}\leq R\}.
$$
\end{assumption}

Such an assumption is widely used in statistical learning theory.
For example, in \cite{Shi2011}, Shi et al. proved that linear
spanning of some smooth kernel functions satisfies Assumption \ref{Assumption1}
with a small $\mu$. By the help of Assumption \ref{Assumption1},
 we can prove that the learning performance of
$L_2$-RBoosting can be essentially improved due to the good
structure of the corresponding estimate.

\begin{theorem}\label{THEOREM3}
Let $0<t<1$, $\mu\in(0,1)$ and  $f_{k}$ be the   estimate defined in
Algorithm \ref{alg2}. If $f_\rho\in \mathcal L_1^r$ and Assumption \ref{Assumption1} holds, then
we have

\begin{equation*}
\begin{aligned}
 & \mathcal E(f_{ k})-\mathcal E(f_\rho)
           \leq  \\
        & C\log\frac2t(3M \! + \! \mathcal
         B)\! ^2
         \!\left(n^{-2r}+k^{-1}+\left(\frac{(kn^{-r}+\sqrt{k})^\mu}{m}\right)^{\frac{2-\mu}{2+\mu}}
        \! \right).
\end{aligned}
\end{equation*}
\end{theorem}

It can be found in Theorem \ref{THEOREM3} that if
$\mu\rightarrow0$, then the learning rate of $L_2$-RBoosting can be
near to $m^{-1}$. This depicts that, with good weak learners,
$L_2$-RBoosting can reach a fairly fast learning rate.

\subsection{Adaptive parameter-selection strategy for $L_2$-RBoosting}

In the previous subsection, we point out that $L_2$-RBoosting is
potentially better than $L_2$-DDRBoosting. In consequence, how to select the
parameter, $u$, is of great importance in  $L_2$-RBoosting. We present an
adaptive    way to fix  the shrinkage degree in this subsection and
show that,  the estimate based on such a parameter-selection
strategy does not degrade the generalization capability very much.
To this end, we split the samples $D_m=(x_i,y_i)_{i=1}^m$ into
two parts of size $[m/2]$ and $m-[m/2]$, respectively (assuming
$m\geq 2$). The first half is denoted by $D_m^l$ (the learning set),
which is used to construct the $L_2$-RBoosting estimate
$f_{D_m^l,\alpha_k,k}$. The second half, denoted by
$D_m^v$ (the validation set), is used to choose $\alpha_k$   by
picking  $\alpha_k\in I:=[0,1]$ to minimize the empirical risk
$$
         \frac{1}{m-[m/2]}\sum_{i=[m/2]+1}^m(y_i-f_{D_m^l,\alpha^*_k,k})^2.
$$
Then, we obtain the estimate
$$
          f^*_{D_m^l,\alpha_k,k}=f_{D_m^l,\alpha^*_k,k}.
$$

Since $y\in[-M,M]$, a straightforward adaptation of
\cite[Th.7.1]{Gyorfi2002} yields that, for any $\delta>0$,
$$
             \mathbf E[\|f^*_{D_m^l,\alpha_k,k}-f_\rho\|_\rho^2]\leq
             1+\delta\inf_{ \alpha_k\in I}\mathbf E[\|f_{D_m^l,\alpha^*_k,k}
             -f_\rho\|_\rho^2]+C\frac{\log m}{m},
$$
holds some positive constant $C$ depending only on $M$, $d$ and
$\delta$. Immediately from Theorem \ref{THEOREM1}, we can conclude:

\begin{theorem}\label{THEOREM4}
Let $f^*_{D_m^l,\alpha_k,k}$ be the adaptive $L_2$-RBoosting
estimate. If $f_\rho\in \mathcal L_1^r$,  then  for arbitrary  constants $k,u\in N$,

\begin{equation*}
\begin{aligned}
           & \mathbf E\left\{ {\mathcal E}({\pi
           _M}f^*_{D_m^l,\alpha_k,k})
           - {\mathcal E}({f_\rho }
            )\right\}  \leq \\ & C (M+\mathcal B)^2\left(2^\frac{3u^2+14u+20}{8u+8}k^{-1}
             +    (m/k)^{-1}\log m+n^{-2r}\right),
\end{aligned}
\end{equation*}
where $C$ is an absolute positive constant.
\end{theorem}

\section{Numerical results}
In this section,  a series of  simulations and real data
experiments will be  carried out to illustrate our
theoretical assertions.
\subsection{Simulation experiments}
In this part,  we first introduce the
simulation settings, including the data sets, weak learners and
experimental environment. Secondly, we analyze the relationship between shrinkage degree and   generalization capability for the proposed $L_2$-RBoosting by means of ideal performance curve. Thirdly, we draw a performance comparison of $L_2$-Boosting, $L_2$-RBoosting and $L_2$-DDRBoosting. The results illustrate that $L_2$-RBoosting with an appropriate shrinkage degree outperforms  other ones, especially for the high dimensional data simulations. Finally, we justify the feasibility of the adaptive  parameter-selection strategy for shrinkage degree
in $L_2$-RBoosting.
\subsubsection{Simulation settings}
In the following simulations, we generate the data from  the following model:
\begin{equation}
Y = m(X)+\sigma\cdot\varepsilon,
\end{equation}
where $\varepsilon$ is standard gaussian noise and independent of
$X$. The noise level $\sigma$ varies among in $\{0, 0.5, 1\}$, and
$X$ is uniformly distributed on $[-2, 2]^d$ with $d\in\{1,2,10\}$. 9
typical regression functions   are considered in this set of
simulations, where these functions are the same as those in section
IV of \cite{Bagirov2010}.

\begin{itemize}

\item $m_1(x)=2*max(1,min(3+2*x,3-8*x)),$ \\


\item
$
m_2(x)=\begin{cases}
10\sqrt { - x} \sin (8\pi x)&  -0.25 \le x < 0,\\
0&\text{else},
\end{cases},
$\\

\item
$
m_3(x)=3*sin(\pi*x/2),
$\\

\item
$
m_4(x_1,x_2)=x_1*sin(x_1^2)-x_2*sin(x_2^2),
$\\

\item
$
m_5(x_1,x_2)=4/(1+4*x_1^2+4*x_2^2),
$\\

\item
$
m_6(x_1,x_2)=6-2*min(3,4*x_1^2+4*|x_2|),
$\\

\item
$
m_7(x_1, \dots , x_{10})= \sum\limits_{j = 1}^{10} {{{( - 1)}^{j - 1}}{x_j}\sin ({x_j}^2)},
$\\


\item
$m_{8}(x_1, \dots , x_{10})=m_6(x_1+ \dots + x_5, x_6+ \dots + x_{10}),
$\\

\item
$
m_{9}(x_1, \dots , x_{10})=m_2(x_1+ \dots + x_{10}).
$\\
\end{itemize}
For  each regression function and each value  of $\sigma \in \{0,
0.5, 1\}$, we first generate a training set of size $m=500$ and an
independent test set , including  $m'=1000$ noiseless observations. We then evaluate the
generalization capability of each boosting algorithm in terms of root mean squared error (RMSE).

It is known that the boosting trees algorithm requires  the specification of two
parameters. One is the number of splits (or the number of nodes)
that are used for fitting each regression tree. The number of leaves equals the number of splits plus one. Specifying $J$ splits
corresponds to an estimate with up to $J$-way interactions. Hastie
et al. \cite{Hastie2001} suggest that $4 \leq J \leq 8$ generally
works well and the estimate is typically not sensitive to the exact choice of $J$ within that range. Thus, in the following simulations, we use the CART \cite{Breiman1984} (with the number of splits $J=4$)
to build up the week learners for regression. Another parameter is
the number of iterations or the number of trees to be fitted. A
suitable value of iterations can range from a few dozen to several thousand, depending on the the shrinkage degree parameter and  which data set we used. Considering the fact that we mainly focus on the impact of the shrinkage degree, the
easiest way to do it is to select the theoretically optimal number
of iterations via the test data set. More precisely, we select   the number
of iterations, $k^*$, as the best one according to $D_{m'}$
directly. Furthermore, for the additional shrinkage degree parameter,
$\alpha_k=2/(k+u), u \in \mathbf{N}$, in $L_2$-RBoosting, we create 20 equally spaced values of $u$ in logarithmic space between $1$ to $10^6$.

All numerical studies are implemented using MATLAB R2014a on a Windows
personal computer with Core(TM) i7-3770 3.40GHz CPUs and RAM 4.00GB,
and the statistics are averaged based on 20 independent trails for
each simulation.

\subsubsection{Relationship between shrinkage degree  and generalization performance }

For each given re-scale factor $u \in [1,10^{6}]$, we employ
$L_2$-RBoosting to train the corresponding estimates on the whole
training samples $D_m$, and then use the independent test samples
$D_{m'}$ to evaluate their generalization performance.

\begin{figure}[H]
\centering
\subfigure{\label{Fig.sub.a}\includegraphics[height=1.7cm,width=2.8cm]{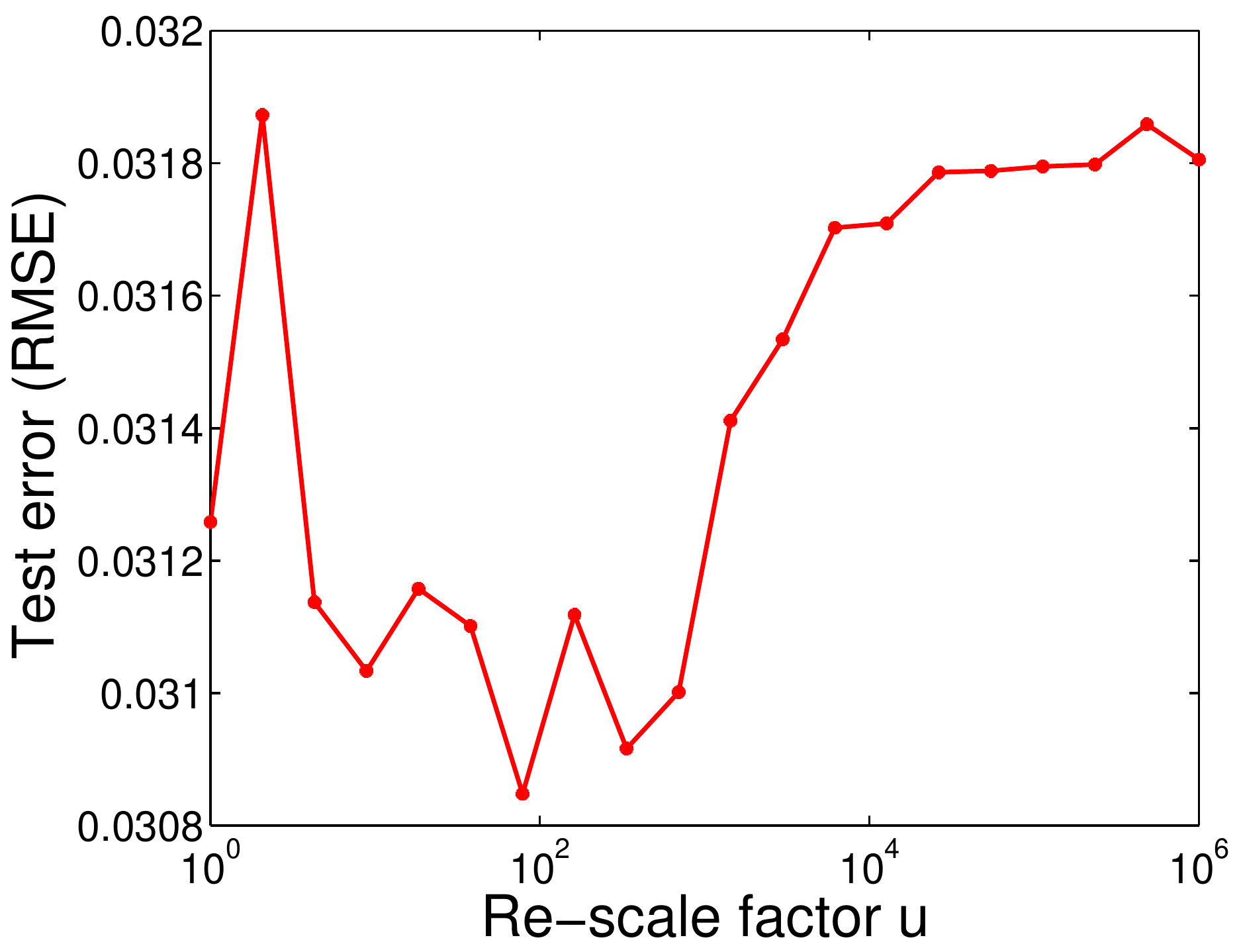}}
\subfigure{\label{Fig.sub.b}\includegraphics[height=1.7cm,width=2.8cm]{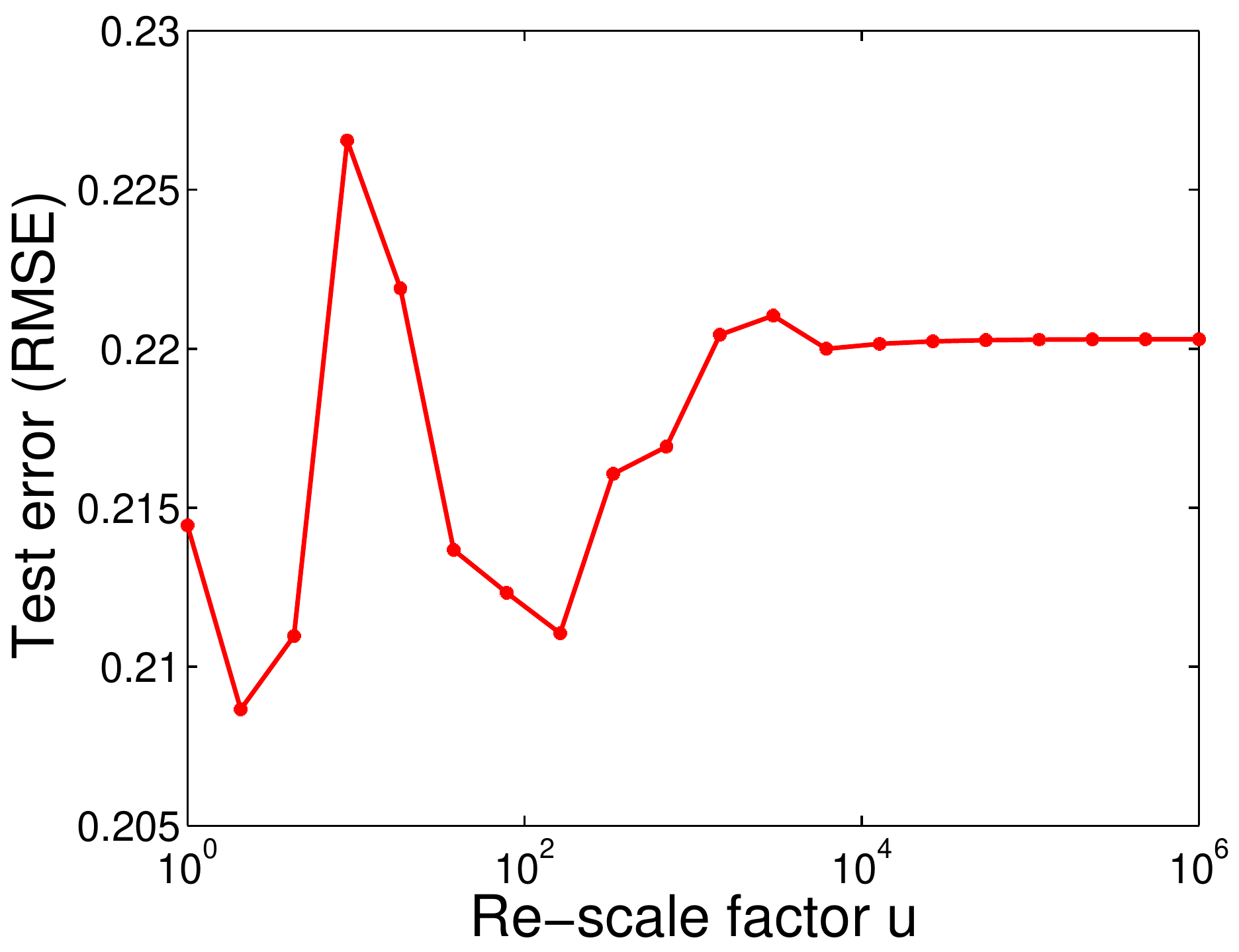}}
\subfigure{\label{Fig.sub.b}\includegraphics[height=1.7cm,width=2.8cm]{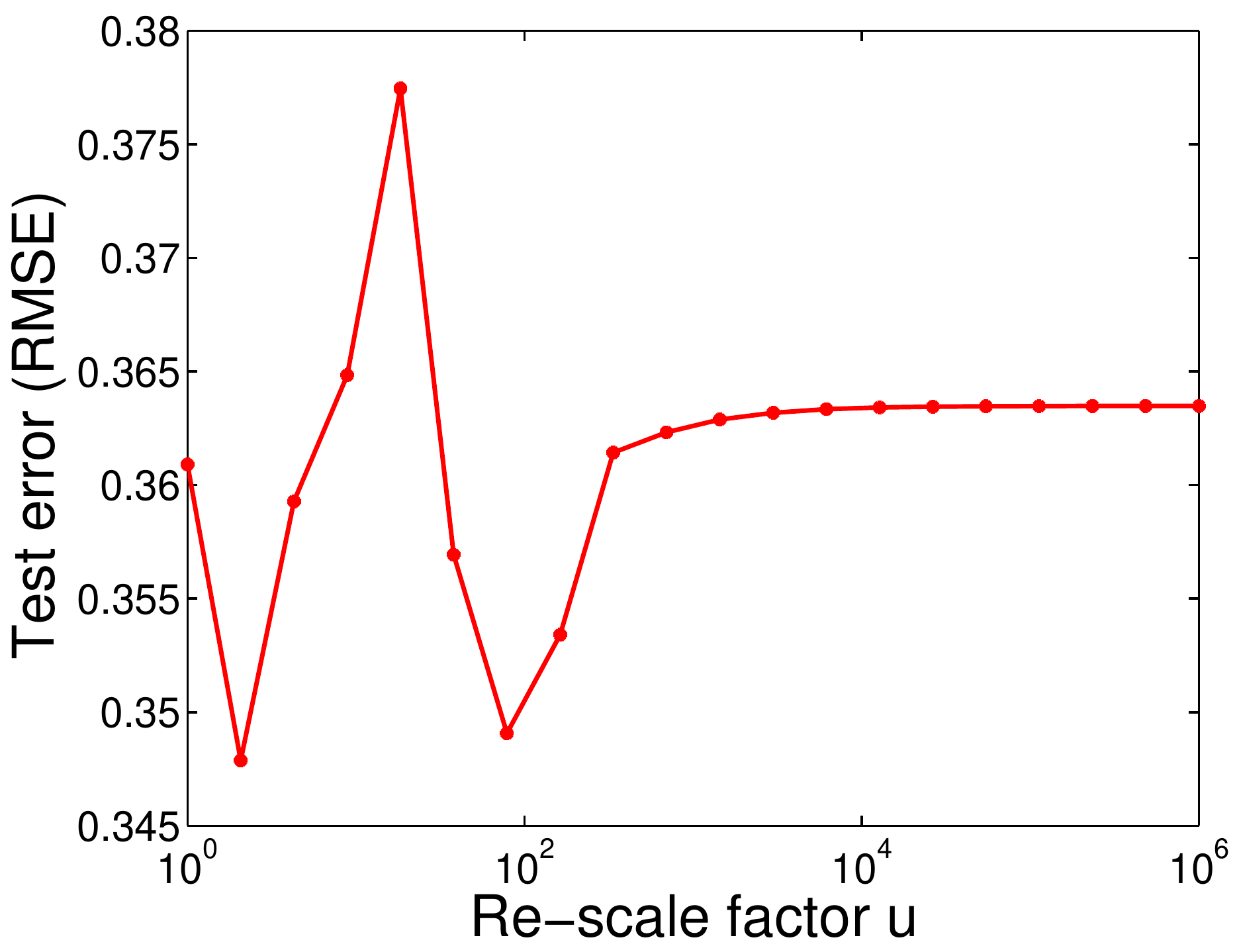}}
%
%
\subfigure{\label{Fig.sub.a}\includegraphics[height=1.7cm,width=2.8cm]{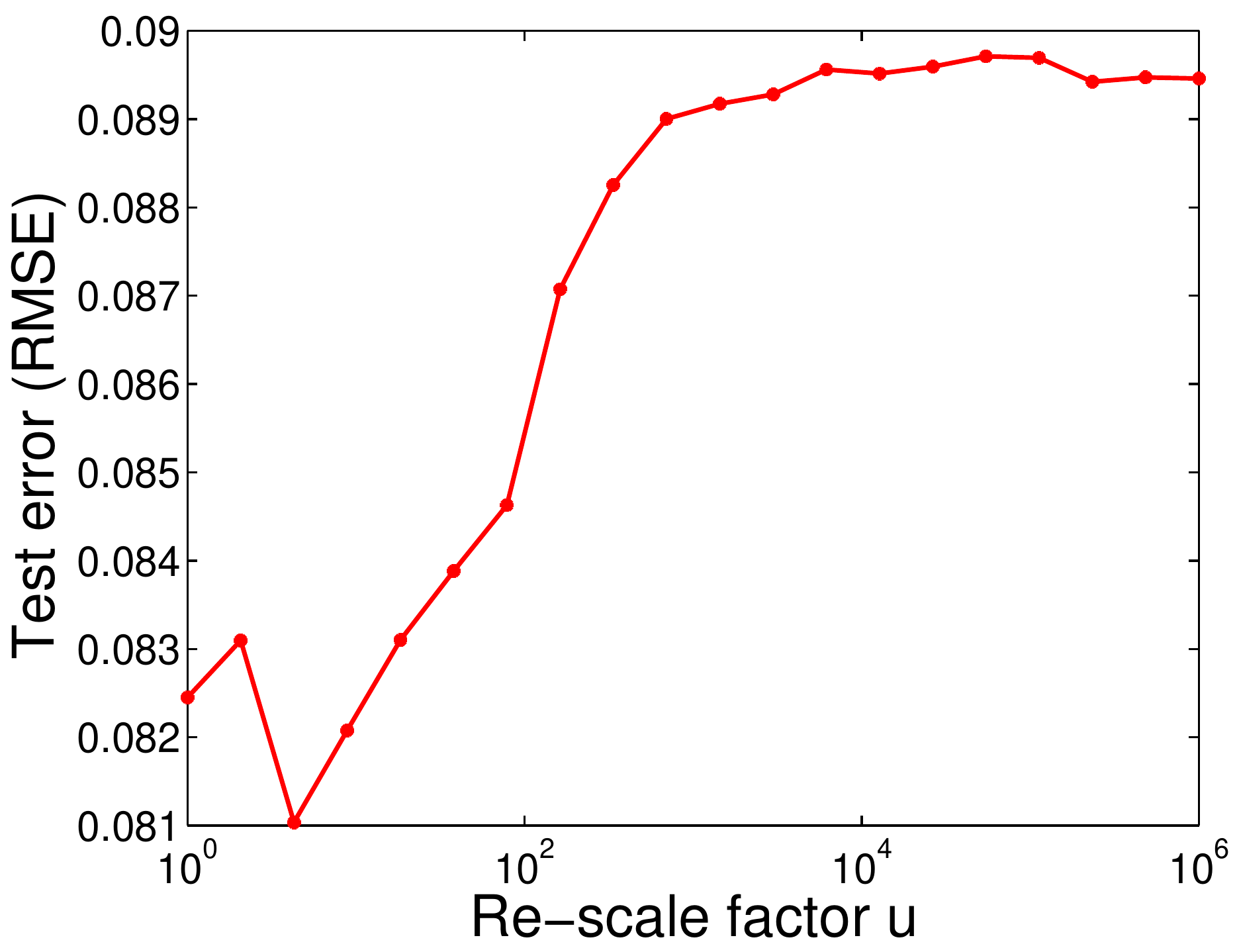}}
\subfigure{\label{Fig.sub.b}\includegraphics[height=1.7cm,width=2.8cm]{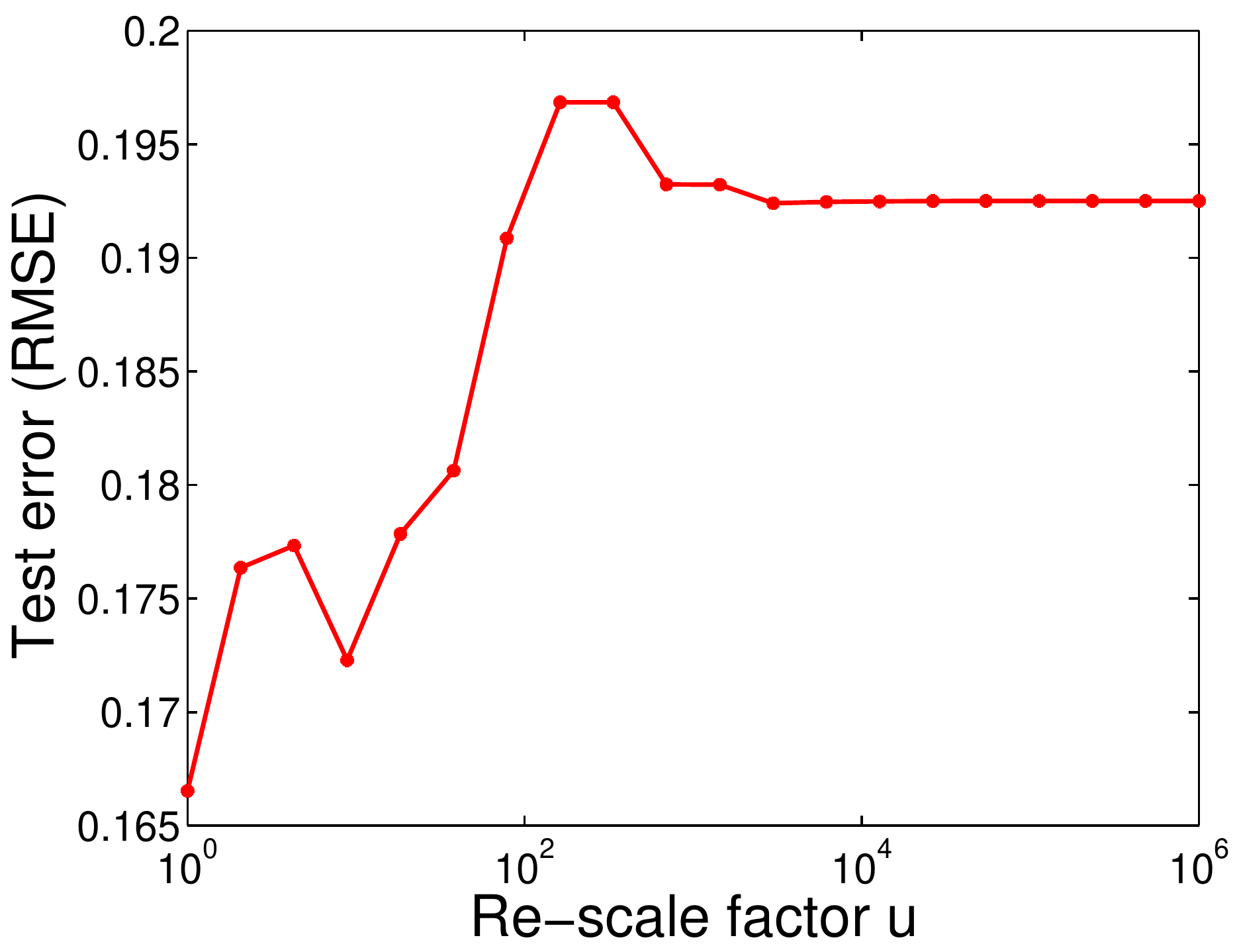}}
\subfigure{\label{Fig.sub.b}\includegraphics[height=1.7cm,width=2.8cm]{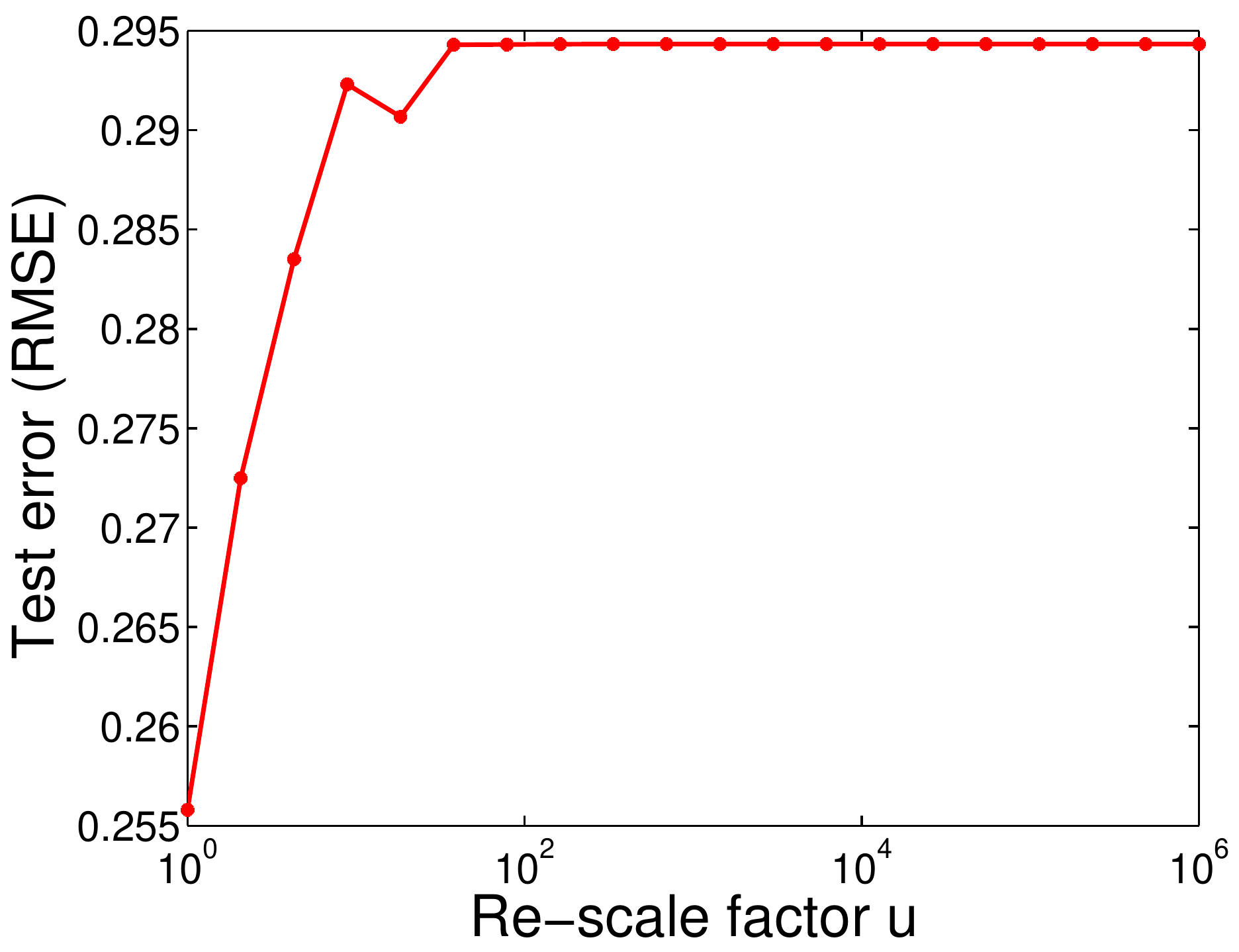}}

\subfigure{\label{Fig.sub.a}\includegraphics[height=1.7cm,width=2.8cm]{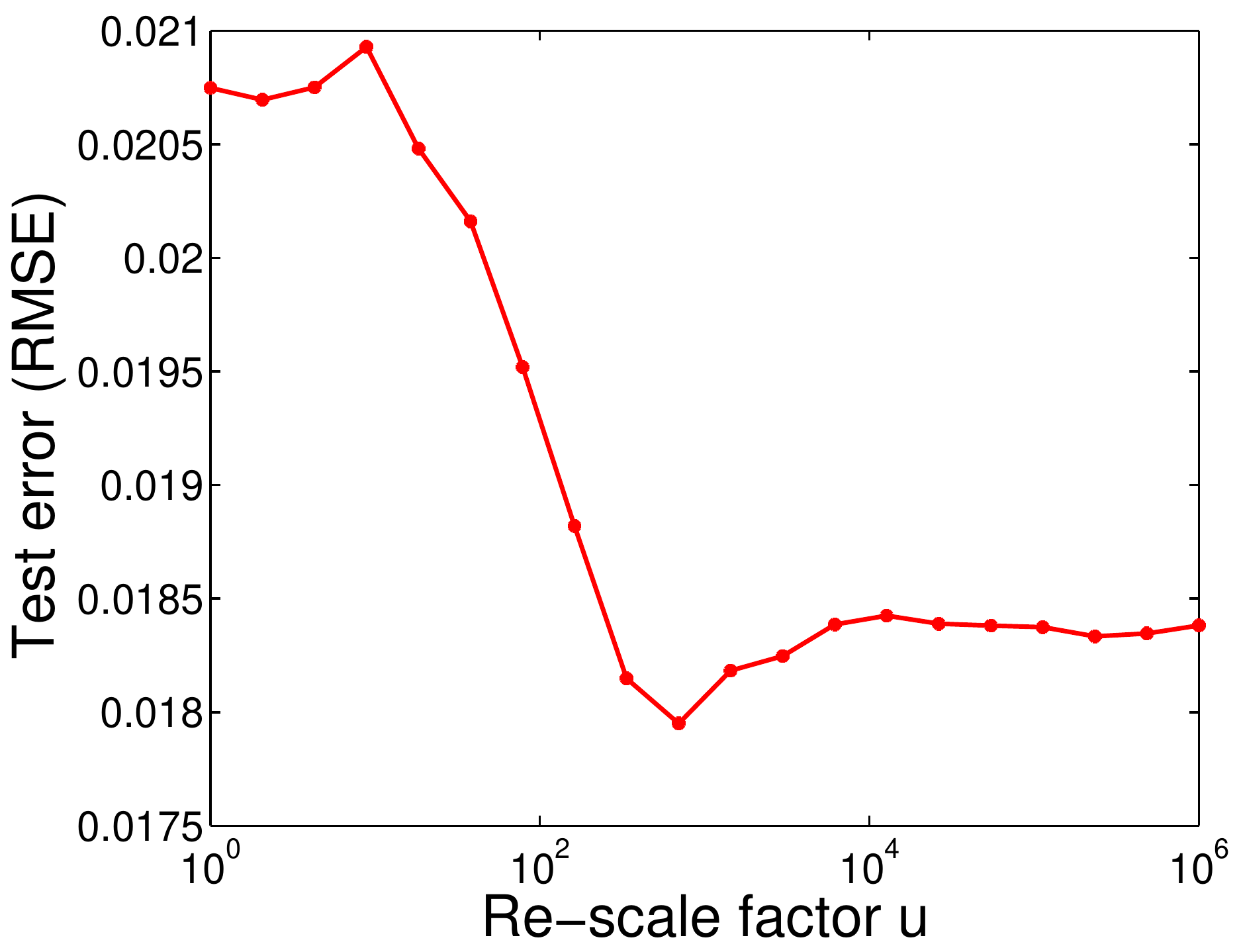}}
\subfigure{\label{Fig.sub.b}\includegraphics[height=1.7cm,width=2.8cm]{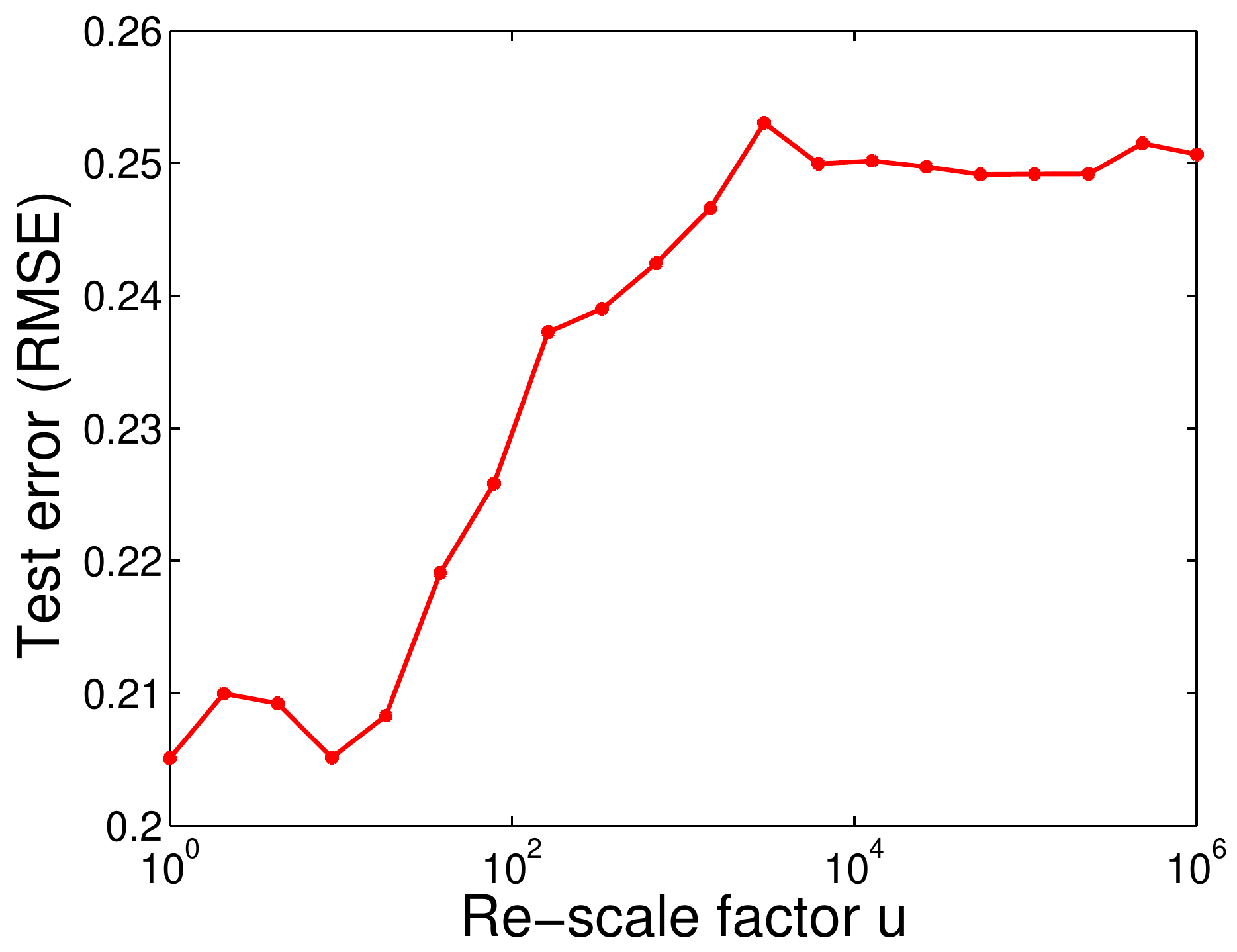}}
\subfigure{\label{Fig.sub.b}\includegraphics[height=1.7cm,width=2.8cm]{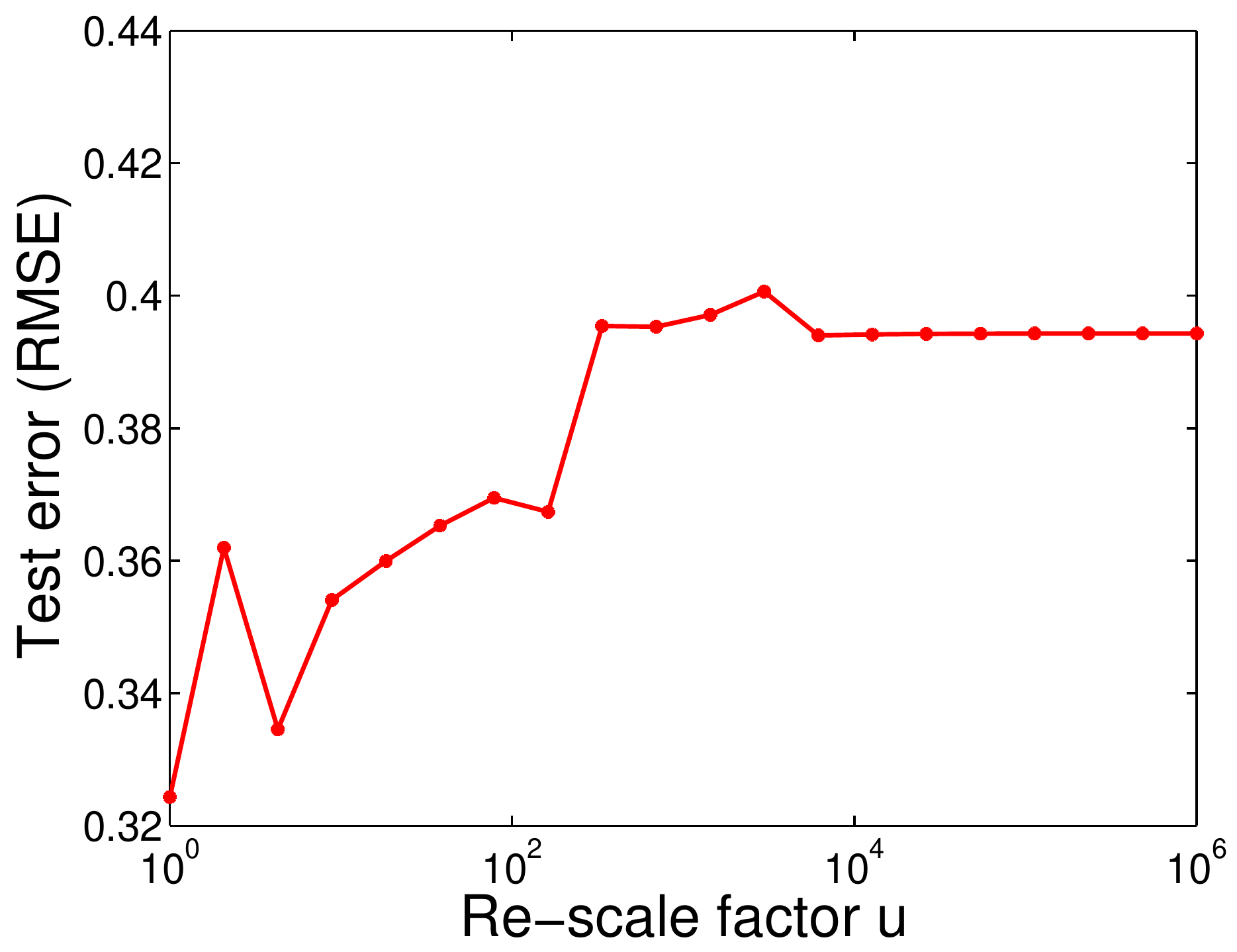}}
\caption{$L_2$-RBoosting  test error (RMSE) curve with respect to
the re-scale factor $u$. Three rows denote the 1-dimension
regression functions $m_1,m_2,m_3$  and three columns indicate the
noise level $\sigma$ varies among in $\{0, 0.5, 1\}$, respectively.}
\label{f1}
\end{figure}

\begin{figure}[H]
\centering
\subfigure{\label{Fig.sub.a}\includegraphics[height=1.7cm,width=2.8cm]{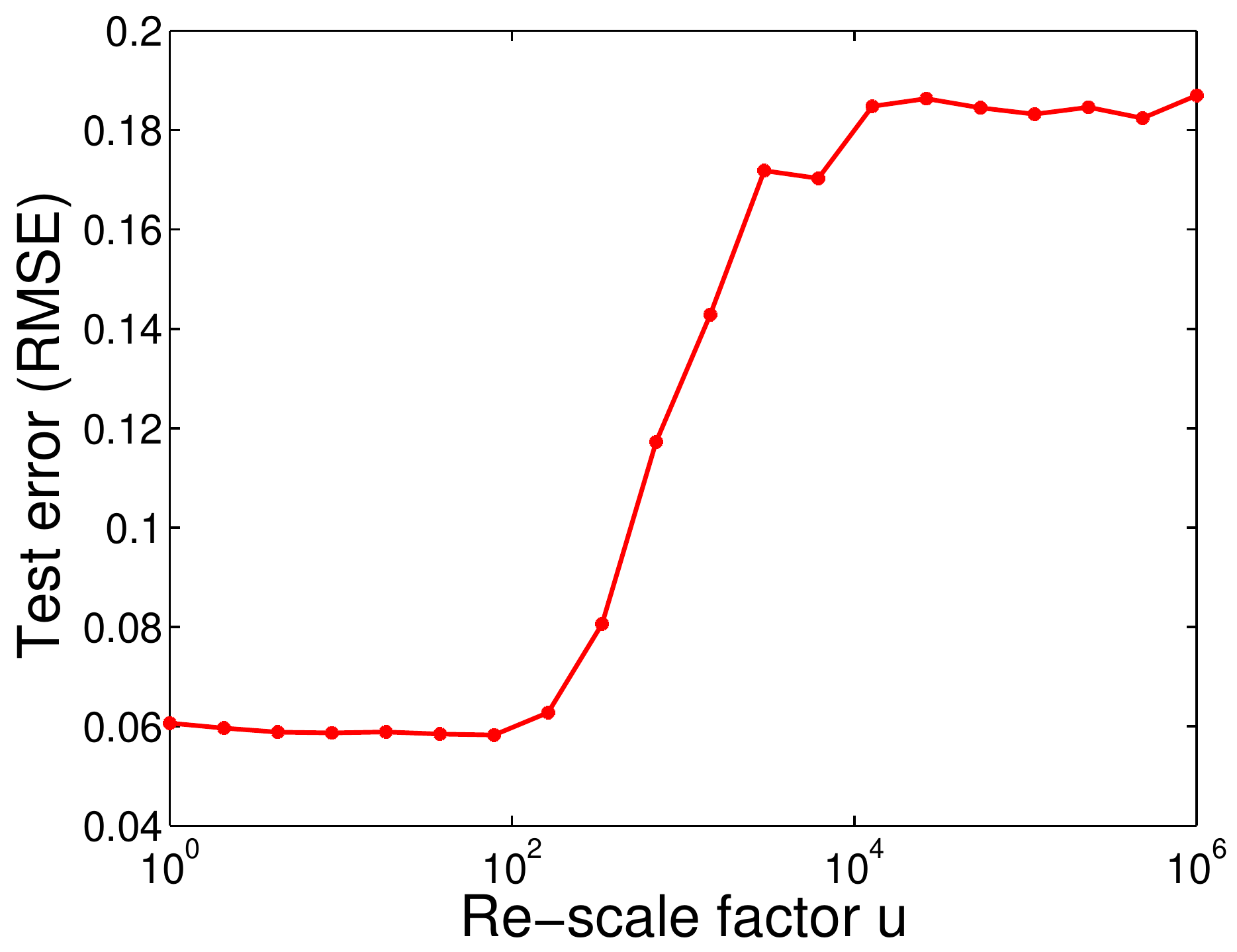}}
\subfigure{\label{Fig.sub.b}\includegraphics[height=1.7cm,width=2.8cm]{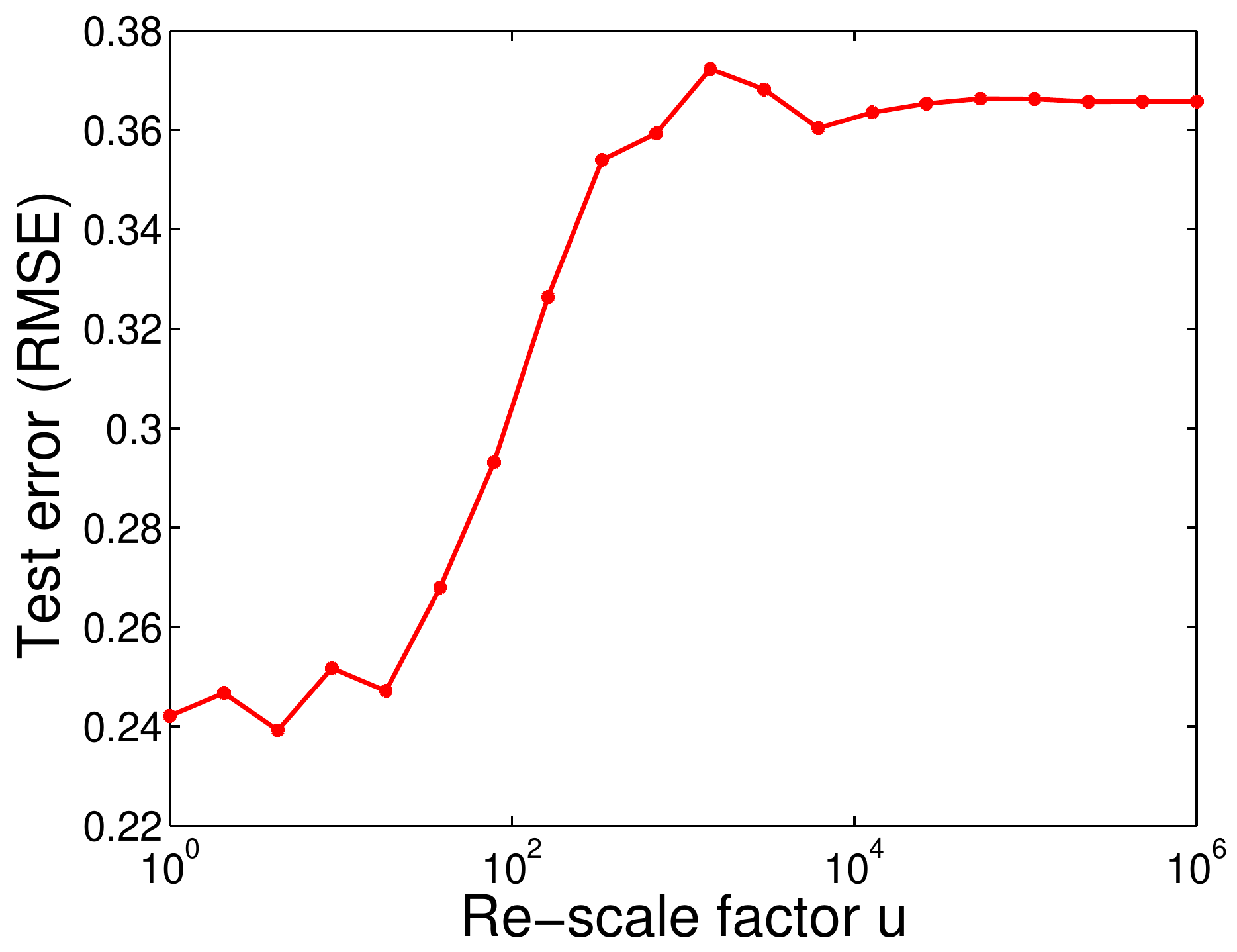}}
\subfigure{\label{Fig.sub.b}\includegraphics[height=1.7cm,width=2.8cm]{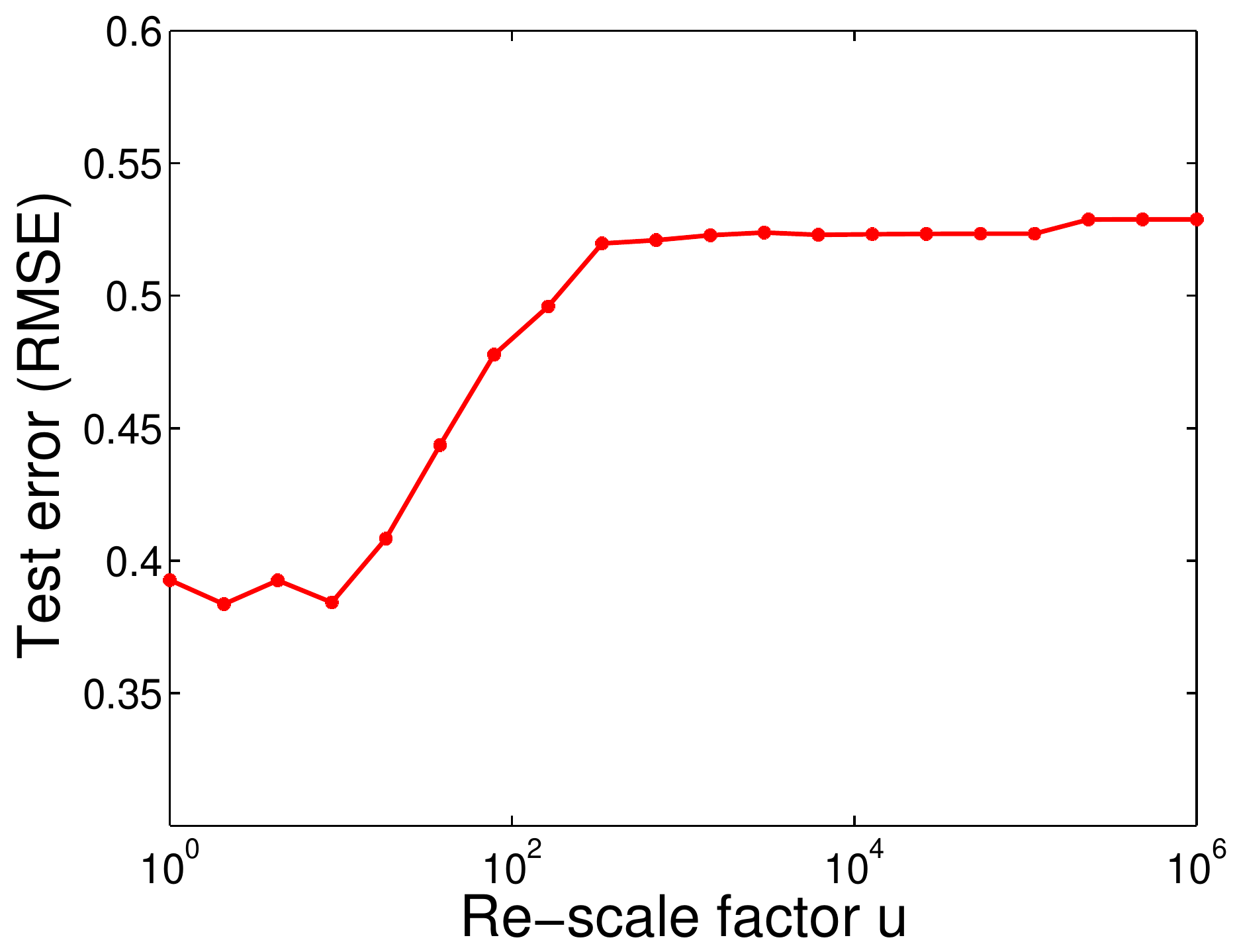}}

\subfigure{\label{Fig.sub.a}\includegraphics[height=1.7cm,width=2.8cm]{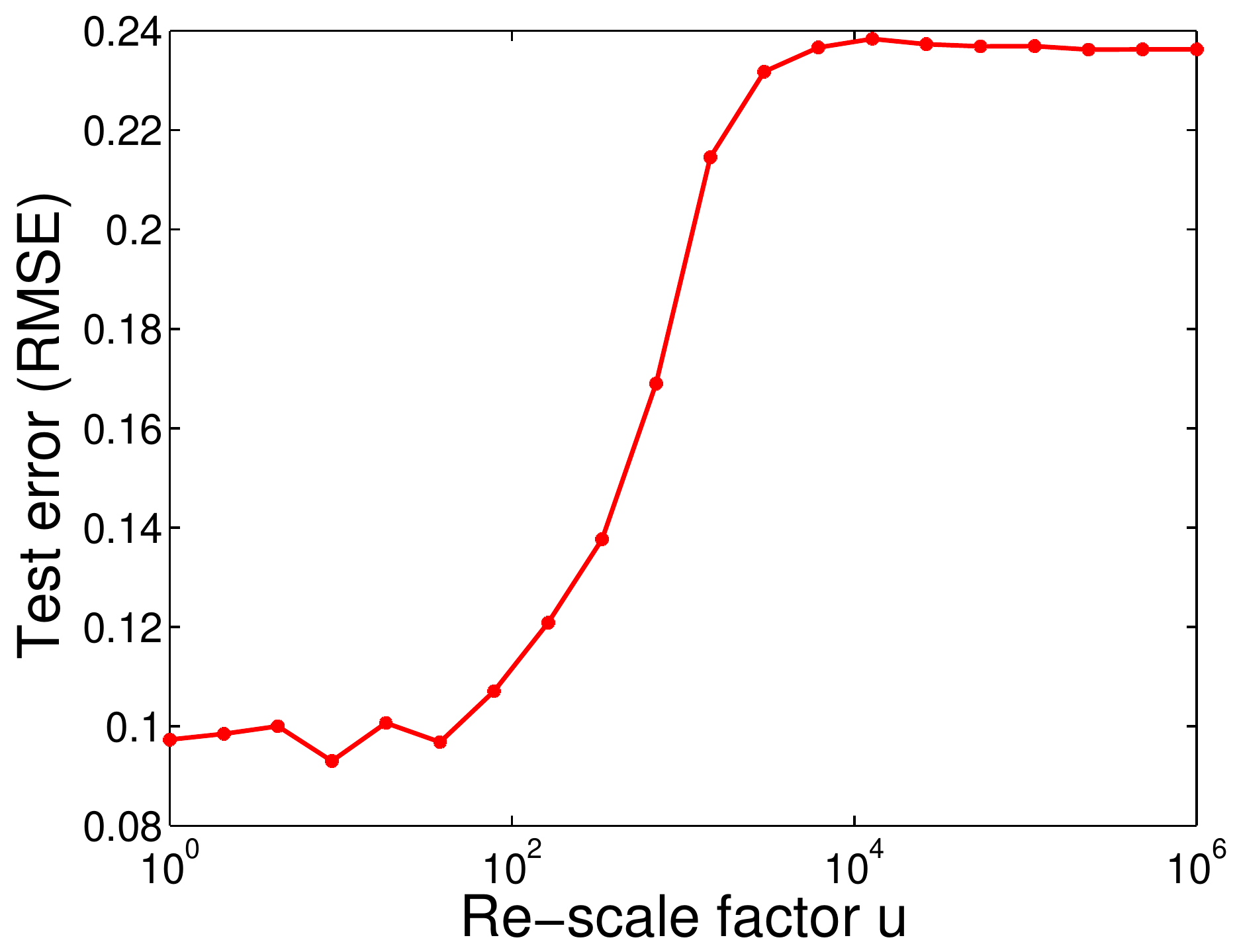}}
\subfigure{\label{Fig.sub.b}\includegraphics[height=1.7cm,width=2.8cm]{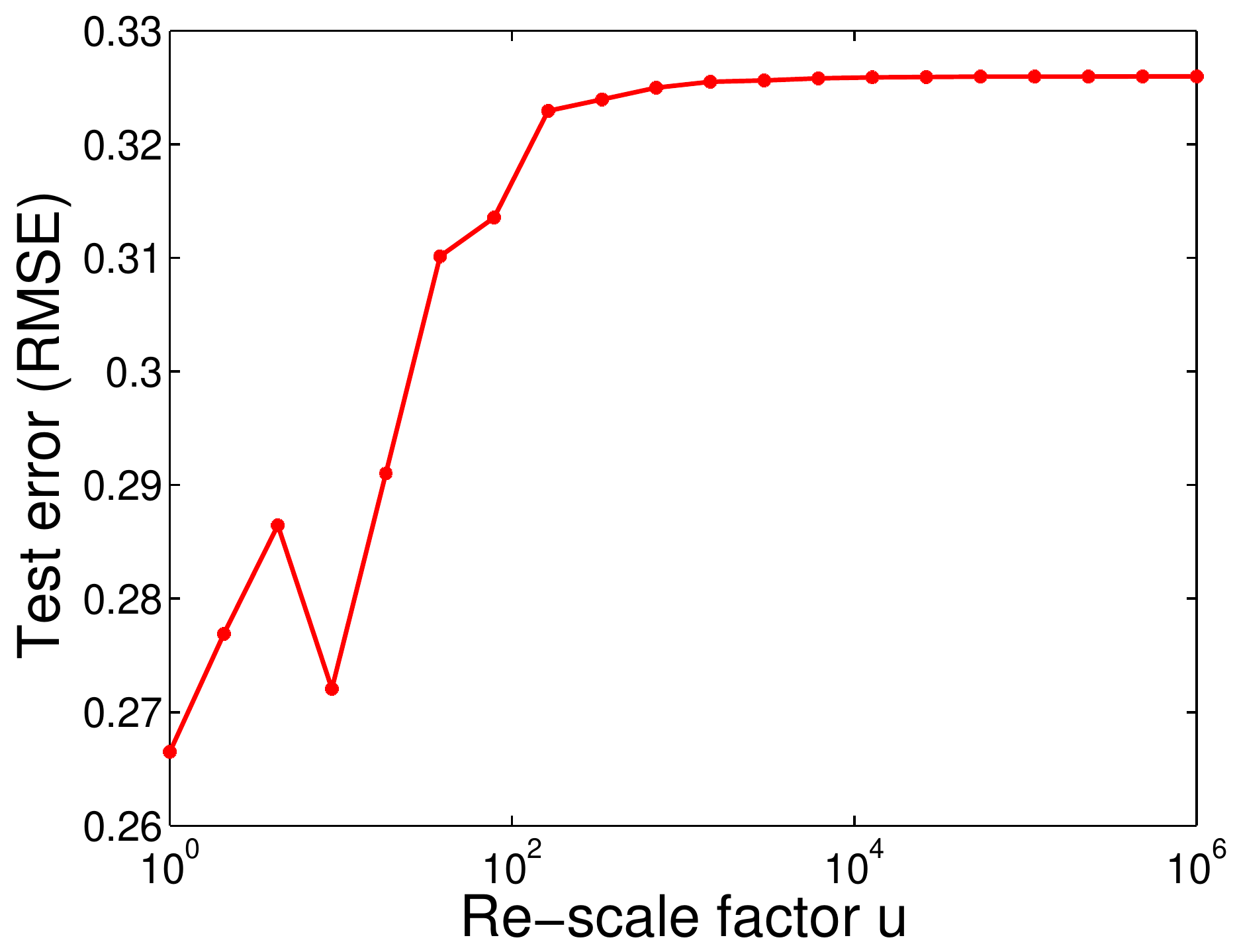}}
\subfigure{\label{Fig.sub.b}\includegraphics[height=1.7cm,width=2.8cm]{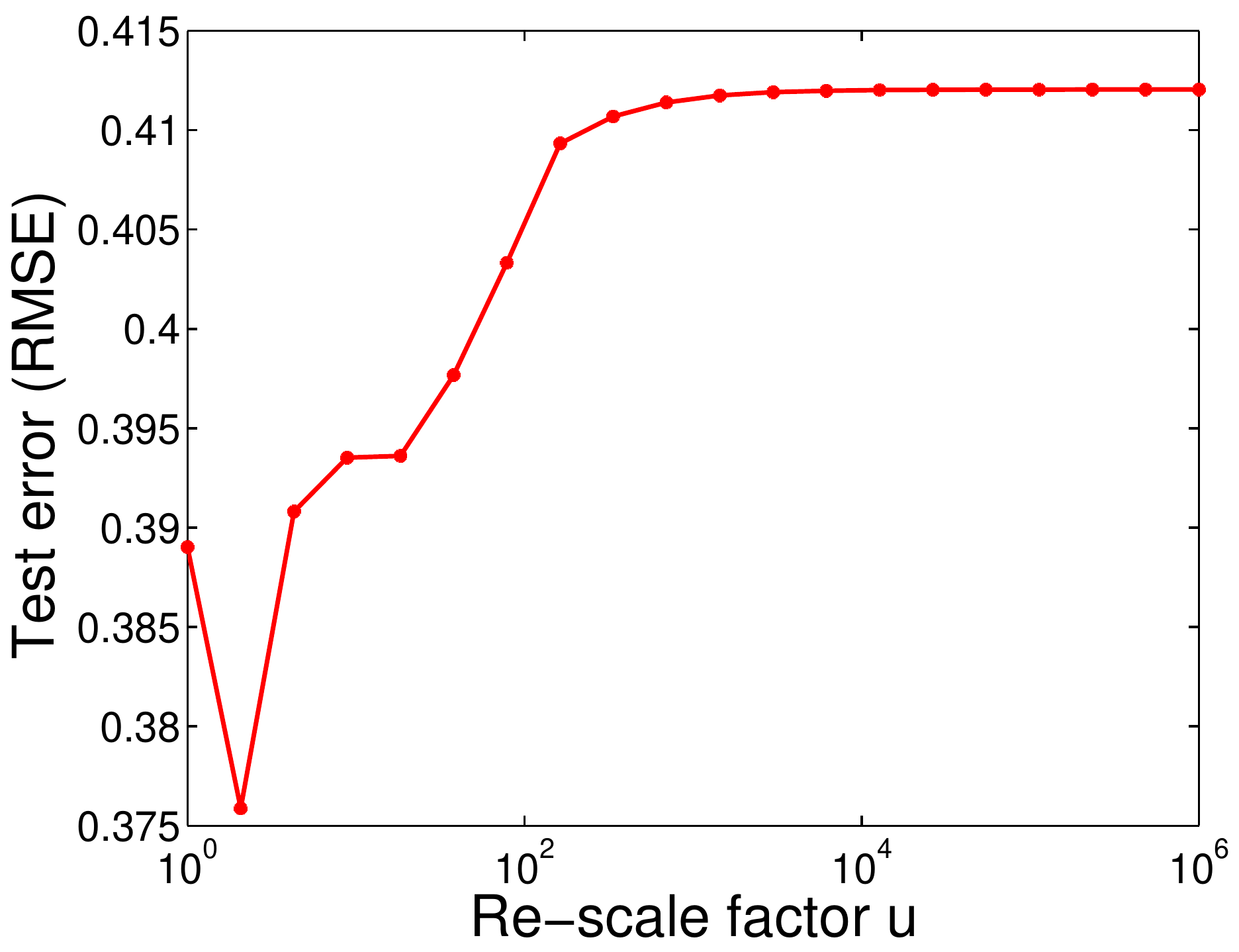}}

\subfigure{\label{Fig.sub.a}\includegraphics[height=1.7cm,width=2.8cm]{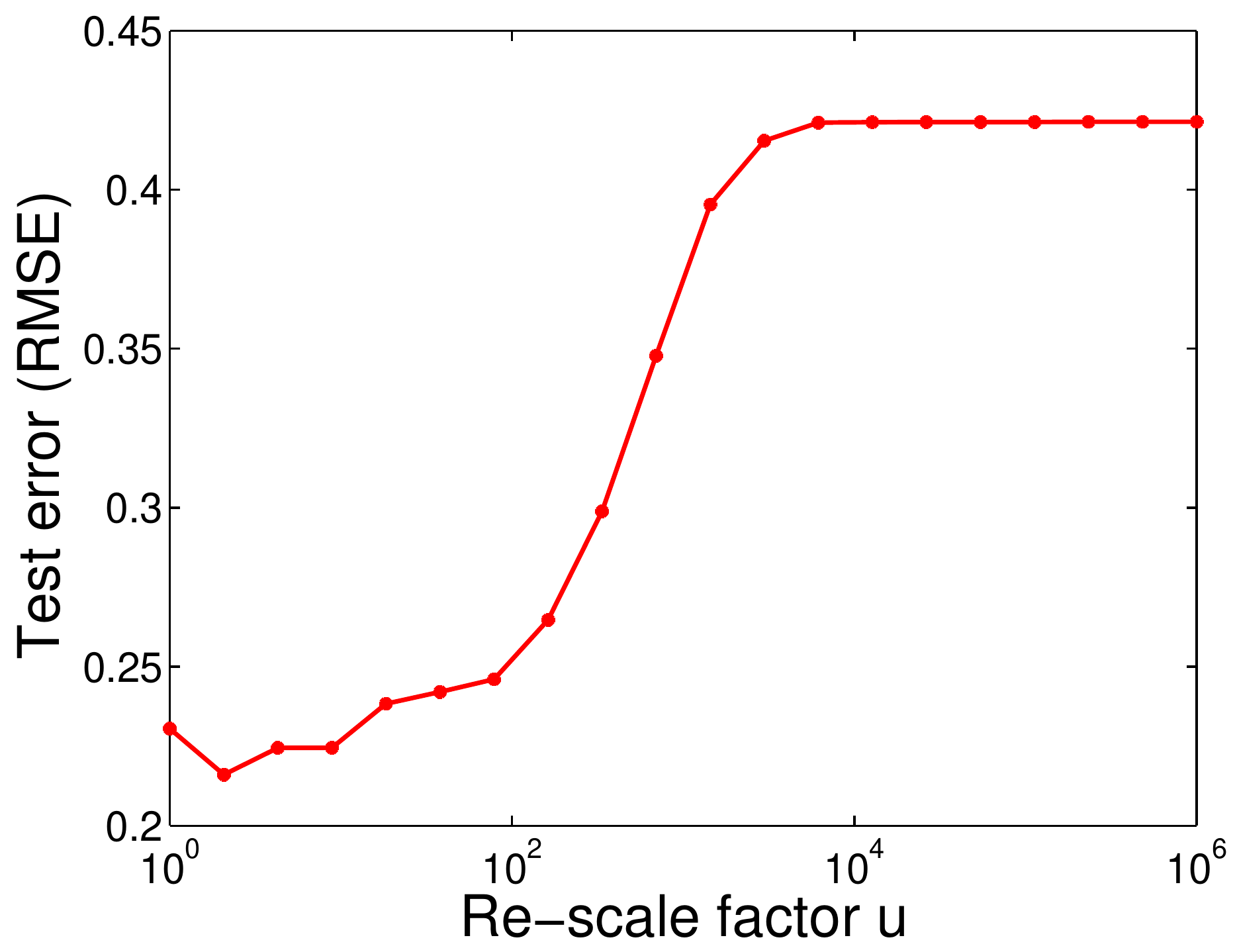}}
\subfigure{\label{Fig.sub.b}\includegraphics[height=1.7cm,width=2.8cm]{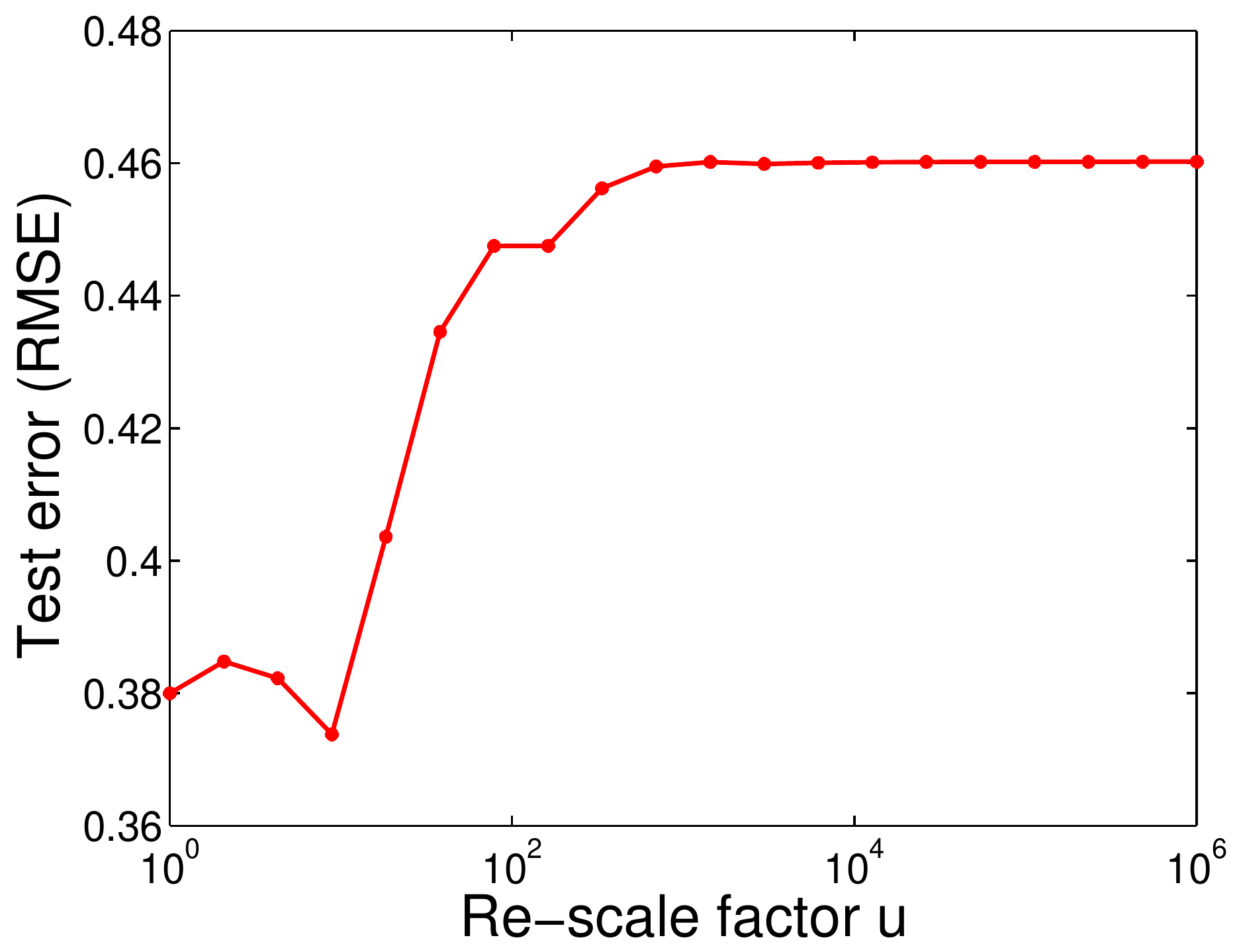}}
\subfigure{\label{Fig.sub.b}\includegraphics[height=1.7cm,width=2.8cm]{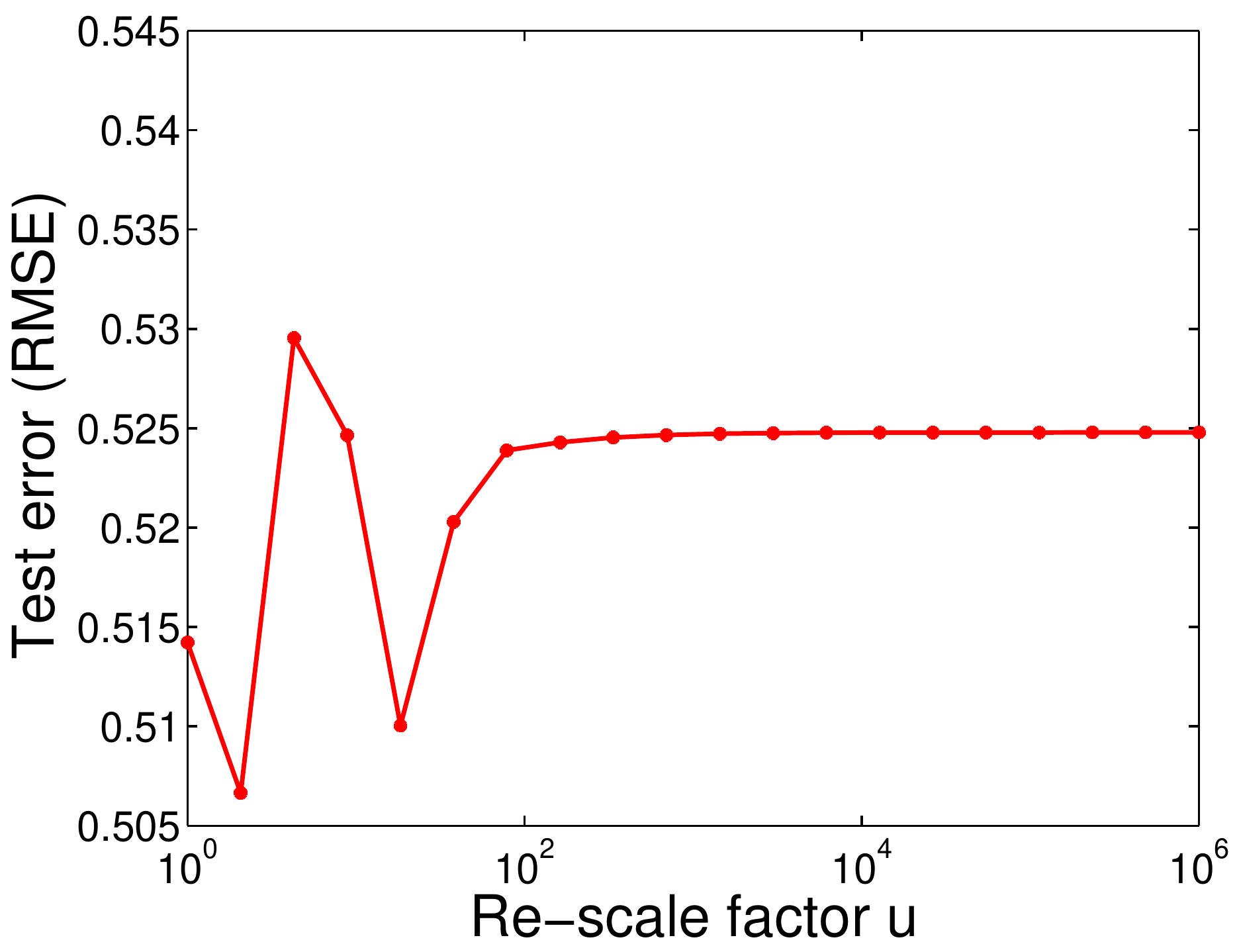}}
\caption{  Three rows denote the 2-dimension regression functions
$m_4,m_5,m_6$  and three columns indicate the noise level $\sigma$
varies among in $\{0, 0.5, 1\}$, respectively. }\label{f2}
\end{figure}

\begin{figure}[H]
\centering
\subfigure{\label{Fig.sub.a}\includegraphics[height=1.7cm,width=2.8cm]{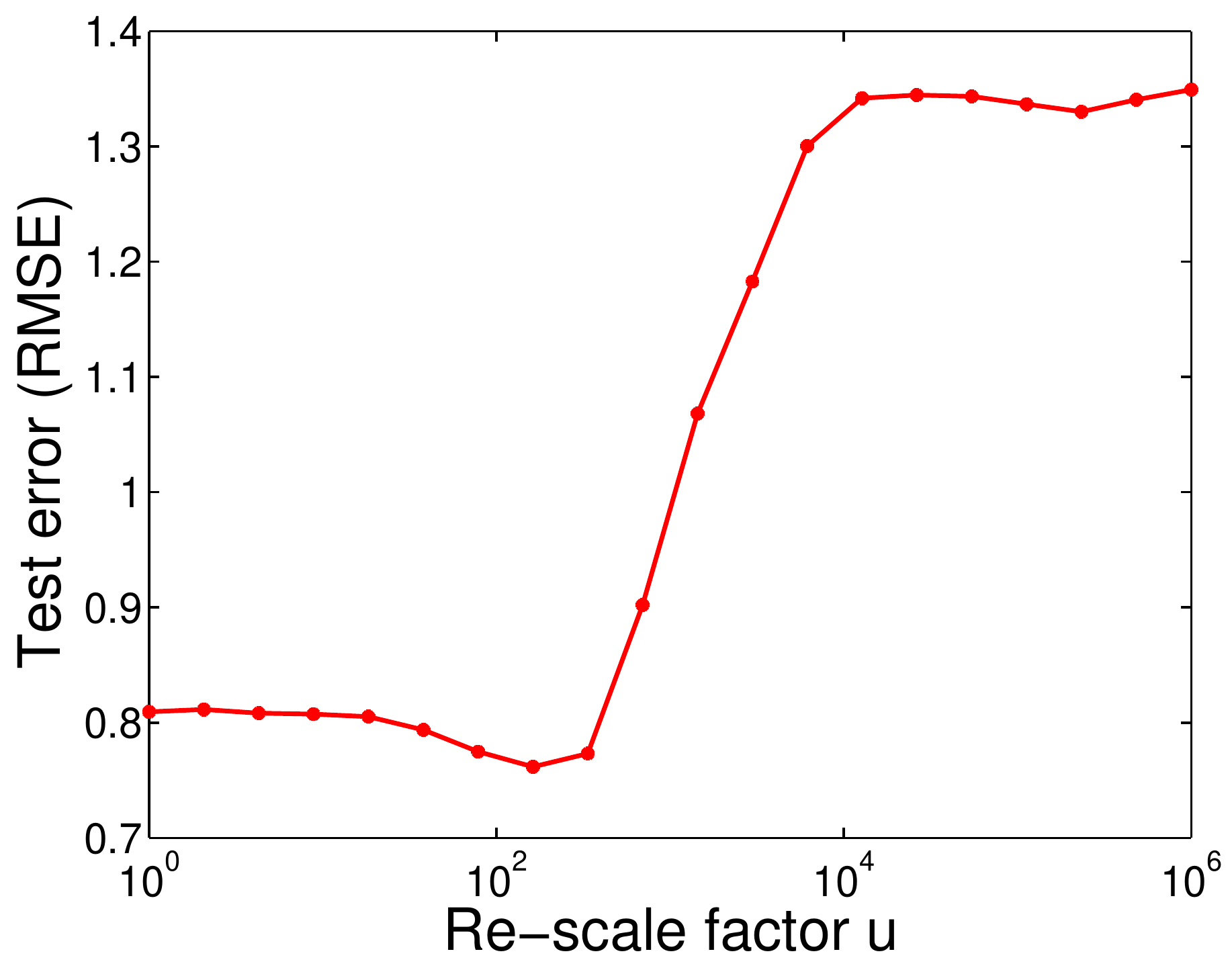}}
\subfigure{\label{Fig.sub.b}\includegraphics[height=1.7cm,width=2.8cm]{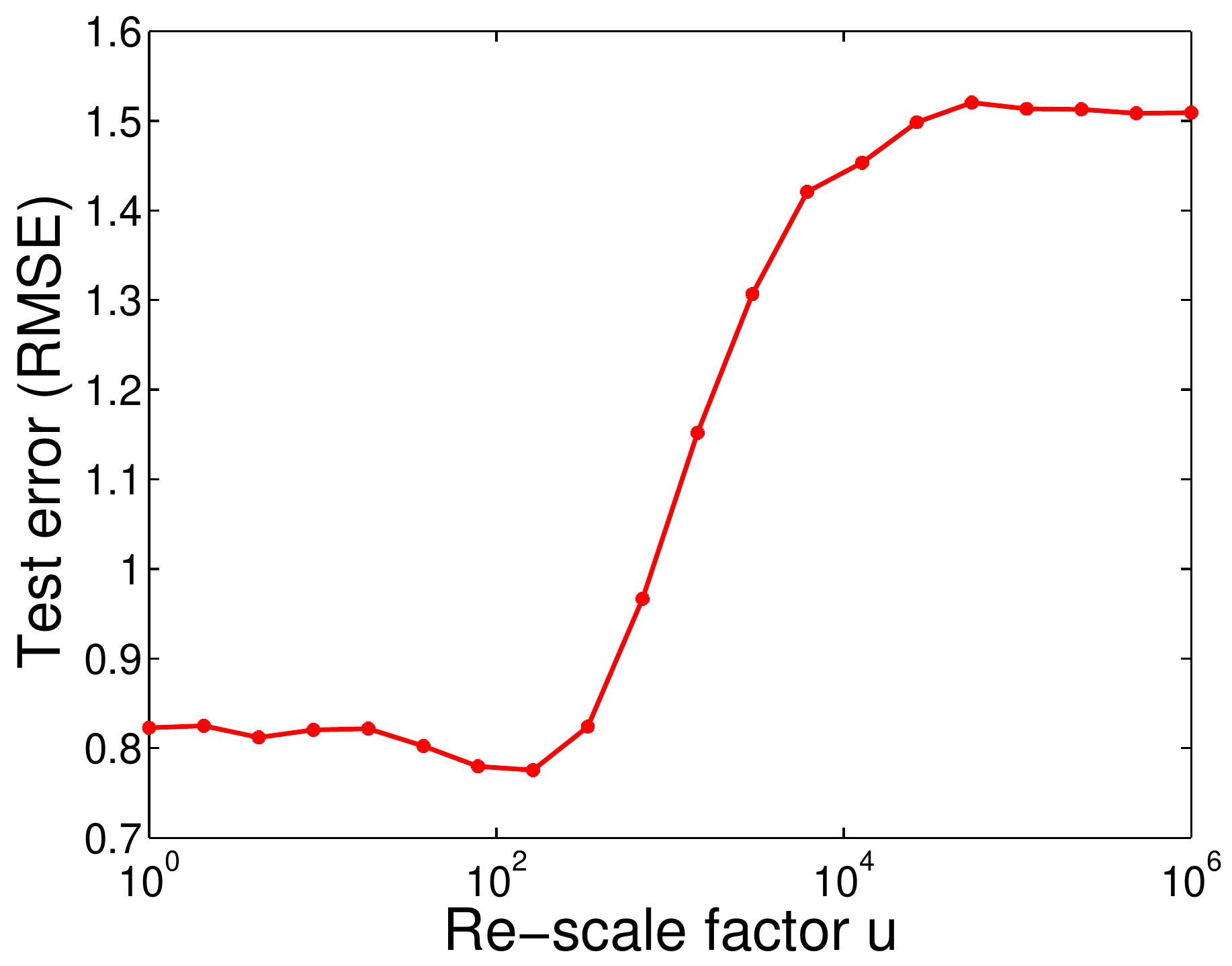}}
\subfigure{\label{Fig.sub.b}\includegraphics[height=1.7cm,width=2.8cm]{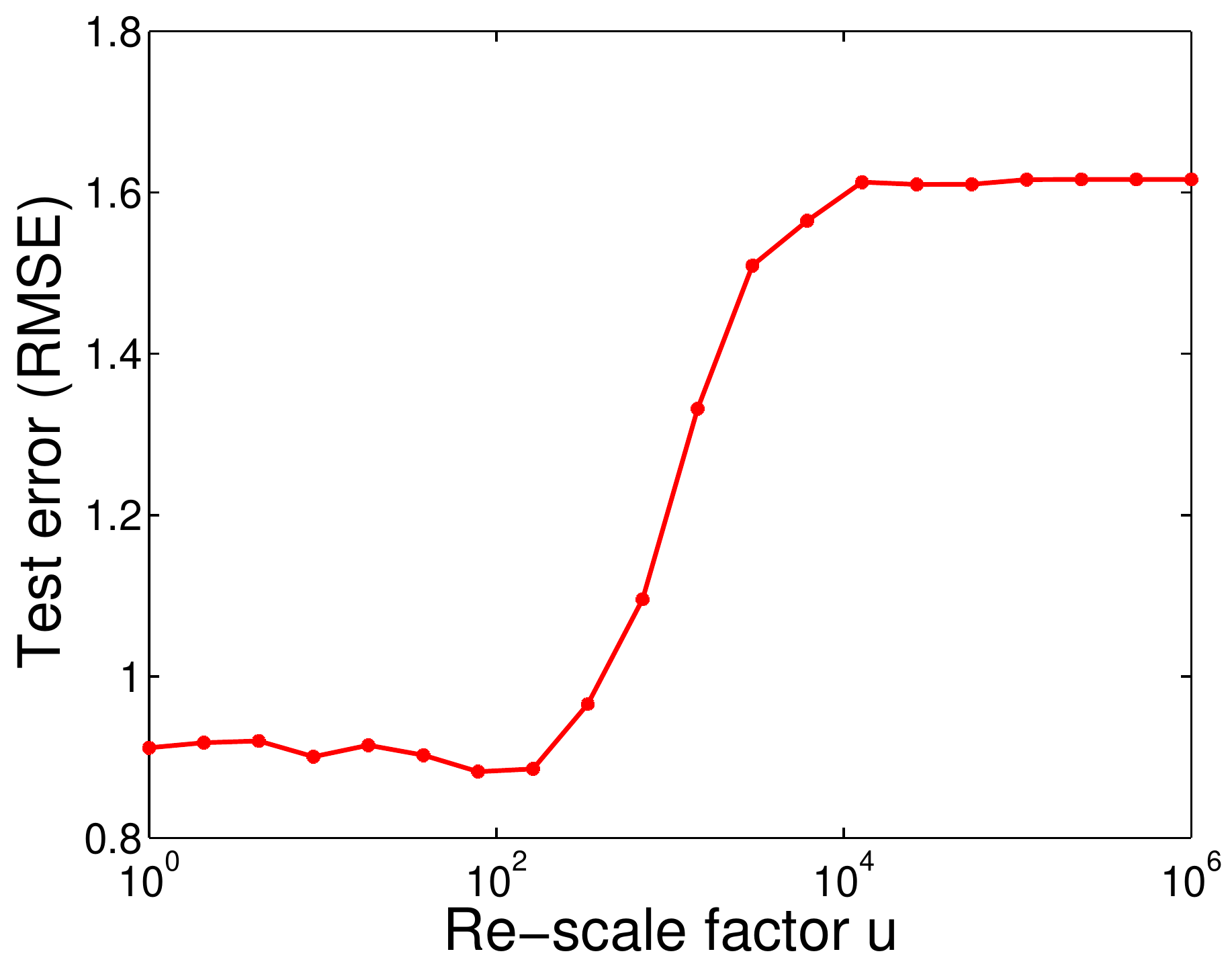}}


\subfigure{\label{Fig.sub.a}\includegraphics[height=1.7cm,width=2.8cm]{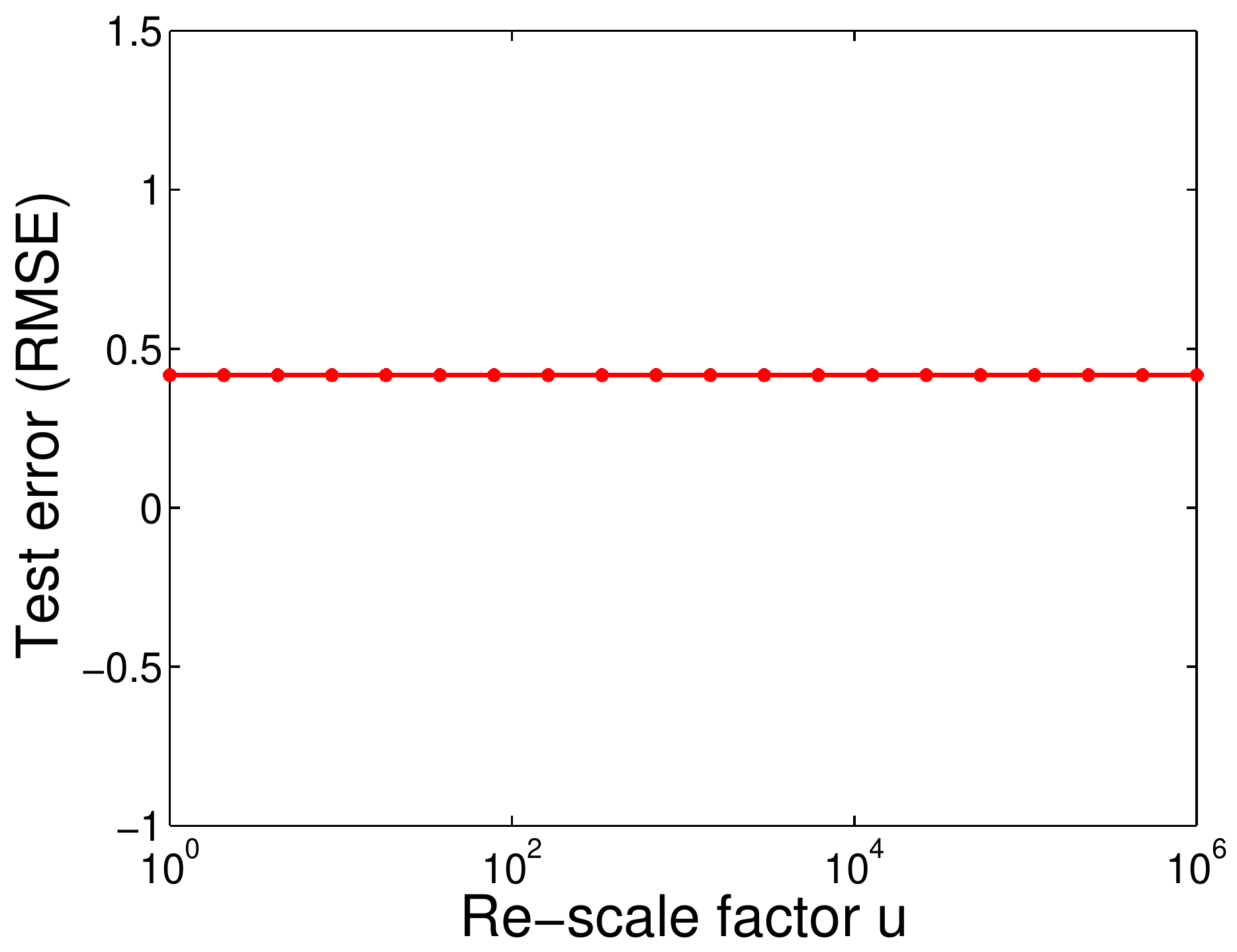}}
\subfigure{\label{Fig.sub.b}\includegraphics[height=1.7cm,width=2.8cm]{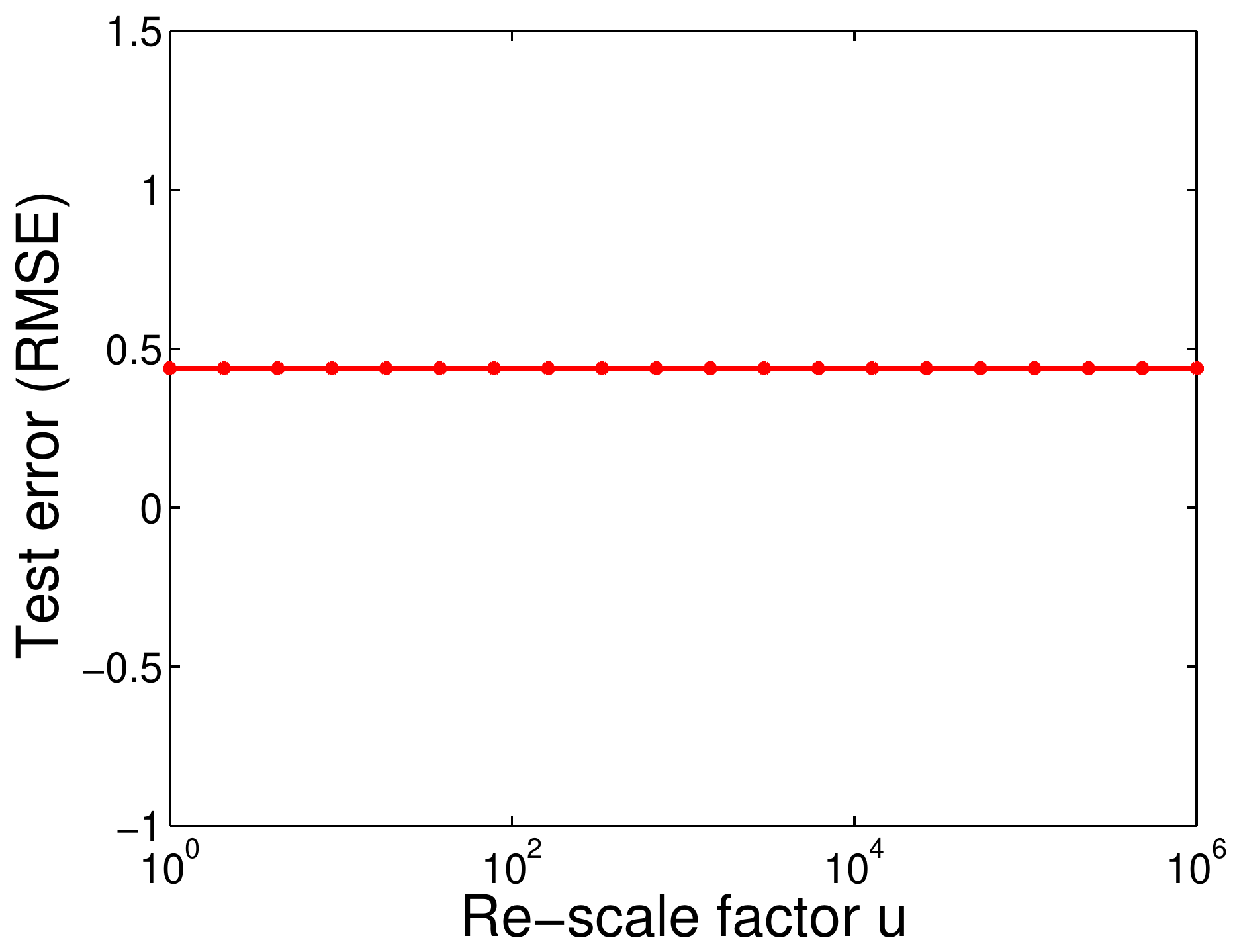}}
\subfigure{\label{Fig.sub.b}\includegraphics[height=1.7cm,width=2.8cm]{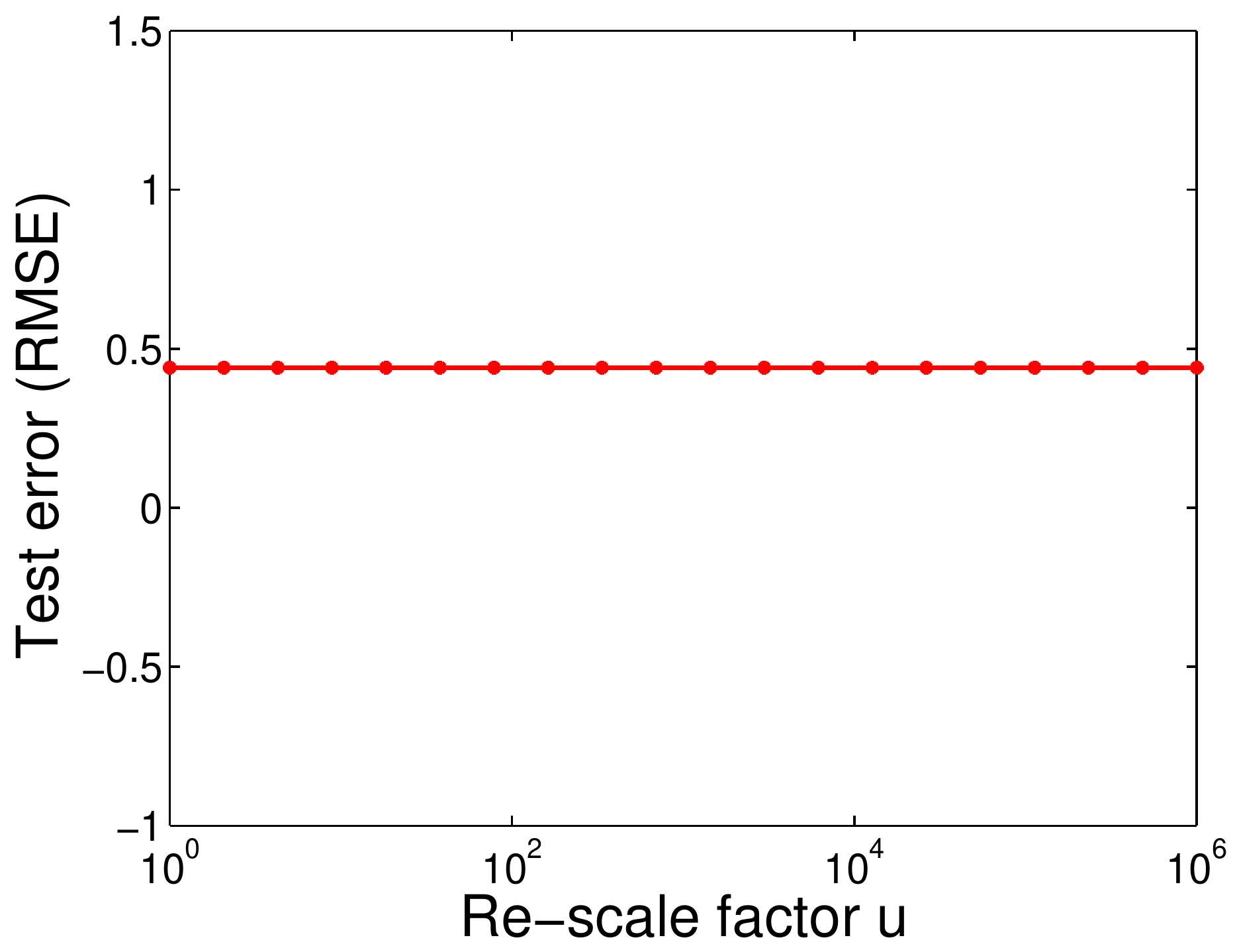}}

\subfigure{\label{Fig.sub.a}\includegraphics[height=1.7cm,width=2.8cm]{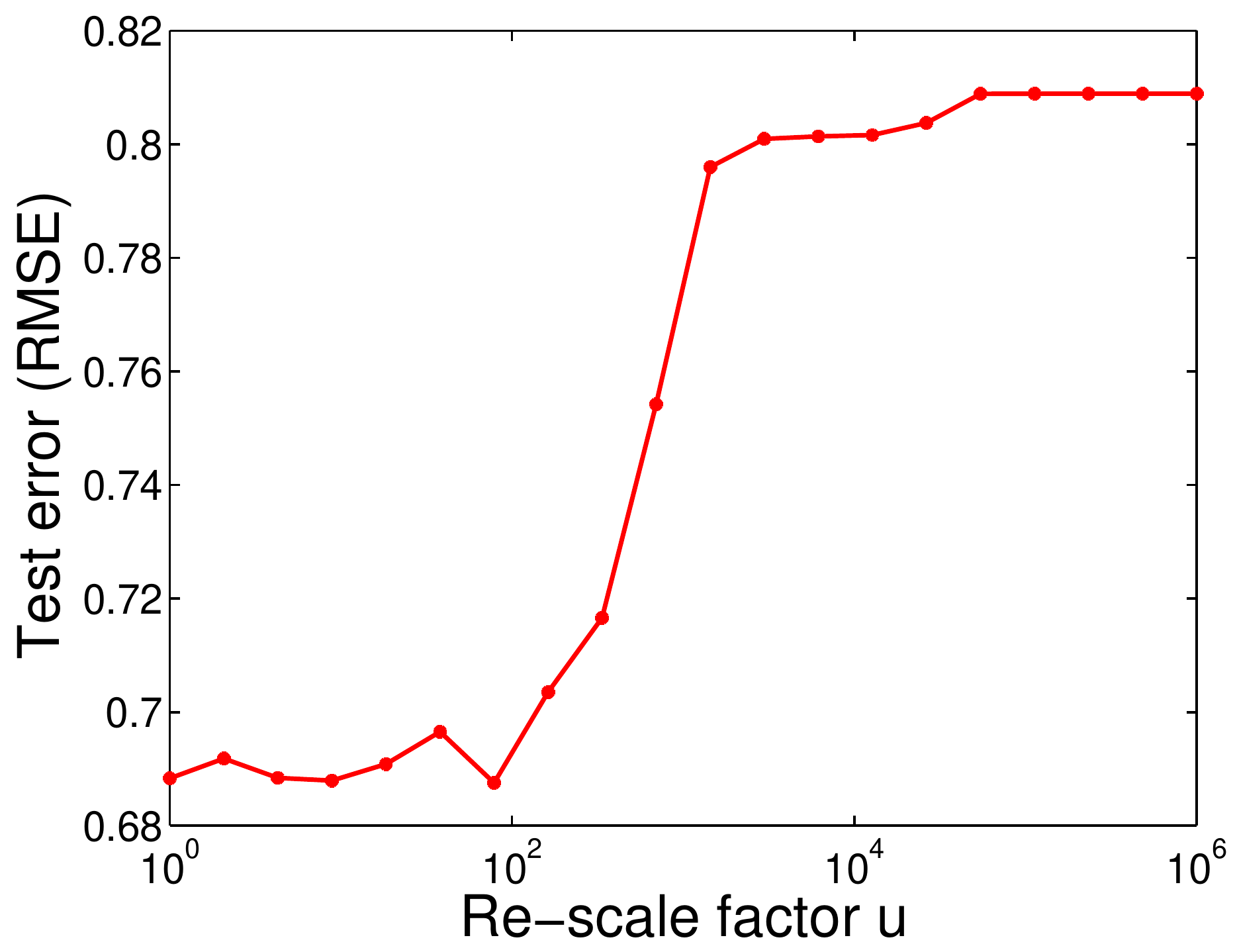}}
\subfigure{\label{Fig.sub.b}\includegraphics[height=1.7cm,width=2.8cm]{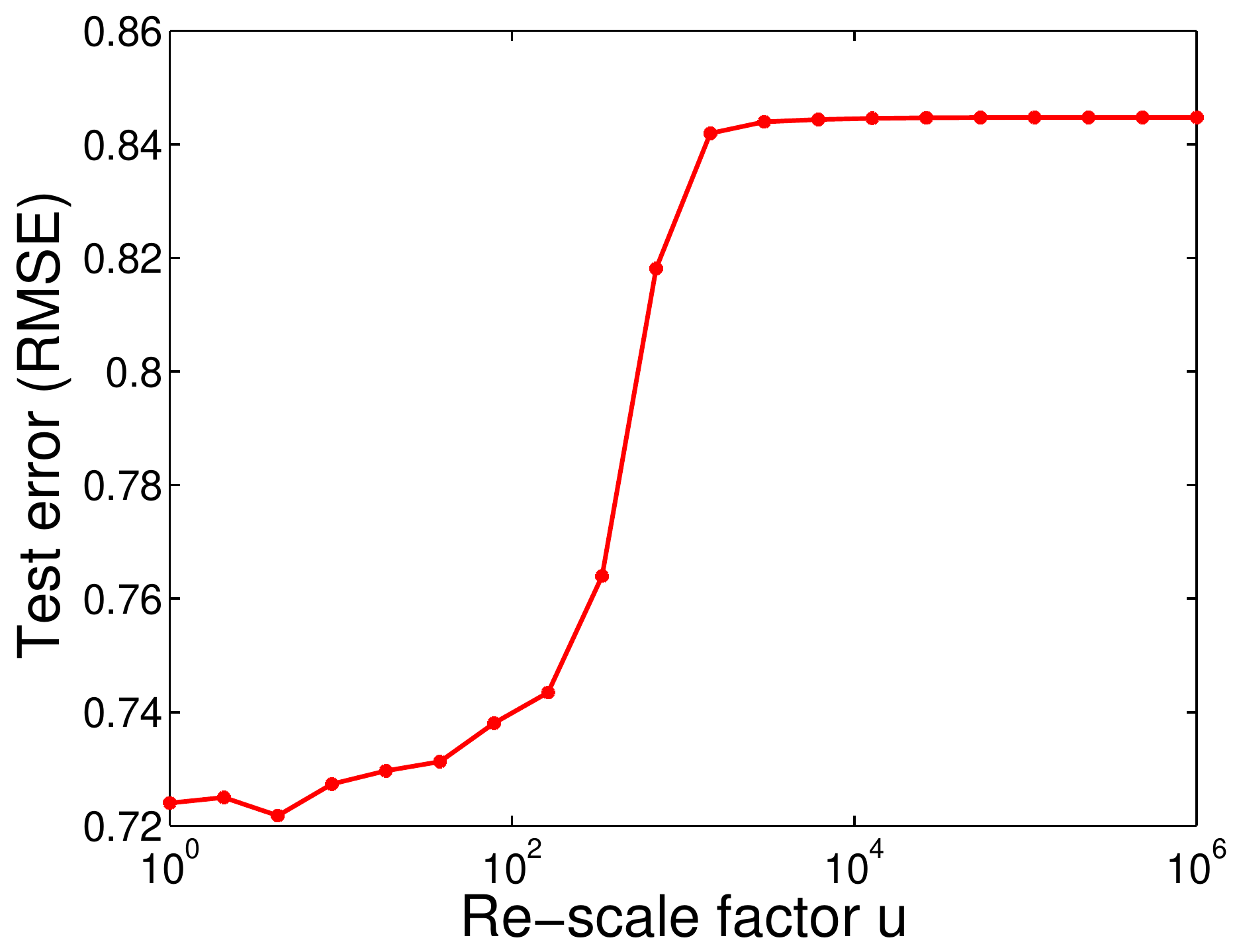}}
\subfigure{\label{Fig.sub.b}\includegraphics[height=1.7cm,width=2.8cm]{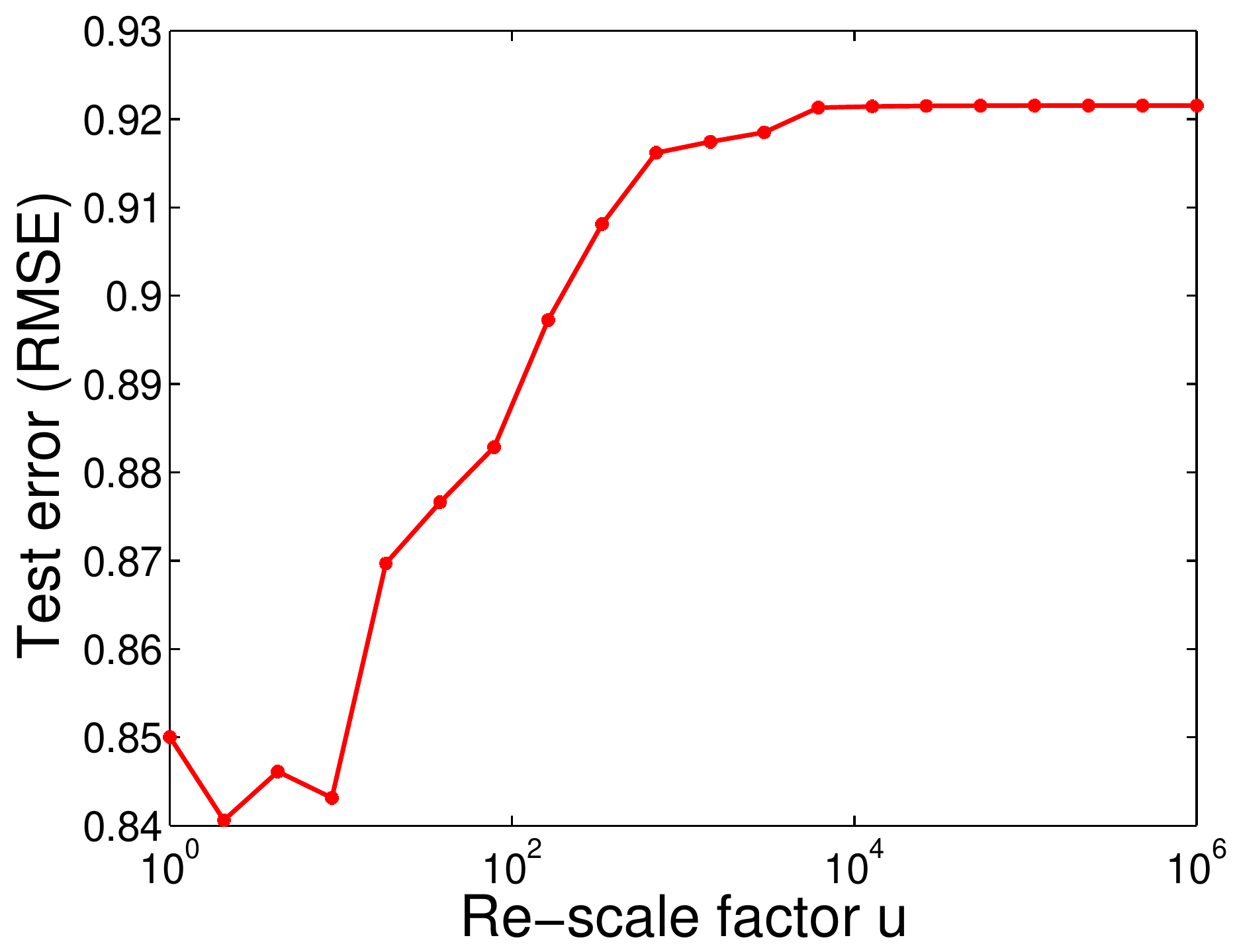}} 
\caption{  Three rows denote the 1-dimension regression functions
$m_7,m_8,m_9$  and three columns indicate the noise level $\sigma$
varies among in $\{0, 0.5, 1\}$, respectively.}\label{f3}
\end{figure}

Fig.\ref{f1}-Fig.\ref{f3} illustrate  the  performance curves of the
$L_2$-RBoosting estimates for the aforementioned nine regression
functions $m_1,\dots,m_9$. It can be easily observed  from these
figures that, except for $m_8$, $u$ has a great influence on the
learning performance of $L_2$-RBoosting. Furthermore, the per-

\noindent
formance curves   generally  imply  that there exists an optimal $u$, which
may be not unique, to minimize the generalization error.
This is consistent with our theoretical assertions.  For $m_8$, the test error curve of $L_2$-RBoosting  is ``flat'' with respect
to $u$, that is, the generalization performance of $L_2$-RBoosting is irrelevant with $u$.
As the uniqueness of the optimal $u$ is not imposed, such numerical
observations do not count our previous theoretical conclusion.
The reason can be concluded as
follows. The first one is that in Theorem \ref{THEOREM1}, we
impose a relatively strong restriction to the regression function
and $m_8$ might be not satisfy it. The other one is that the adopted
weak learner is too strong (we pre-set the number of splits $J=4$).
Over grown tree trained on all samples are
liable to autocracy and re-scale operation does not bring  performance benefits at all in such case.
All these numerical results illustrate the importance of selecting an appropriate shrinkage degree in $L_2$-RBoosting.

\subsubsection{Performance comparison of $L_2$-Boosting, $L_2$-RBoosting and $L_2$-DDRBoosting}


In this part, we compare the learning performances among
$L_2$-Boosting, $L_2$-RBoosting and $L_2$-DDRBoosting.
Table \ref{t1}-Table \ref{t3} document  the generalization errors (RMSE)
of $L_2$-Boosting, $L_2$-RBoosting and $L_2$-DDRBoosting  for
regression functions $m_1, \dots,  m_9$, respectively (the bold
numbers denote the optimal performance). The standard errors are
also reported (numbers in parentheses).

Form the tables we can  get clear results that except for the
noiseless 1-dimensional cases, the performance of $L_2$-RBoosting
dominates both $L_2$-Boosting and $L_2$-DDRBoosting for all
regression functions  by a large margin. Through this  series of
numerical studies, including 27 different learning tasks, firstly,
we verify the second guidance deduced from Thm.\ref{t1} that
$L_2$-RBoosting outperforms $L_2$-Boosting with finite sample
available. Secondly, although $L_2$-DDRBoosting can perfectly solve
the parameter selection problem in the re-scale-type boosting
algorithm, the table results also illustrate that $L_2$-RBoosting
endows better performance once an appropriate $u$ is selected.

%

\subsubsection{Adaptive parameter-selection strategy for shrinkage degree}
We employ the simulations to verify the feasibility
of the  proposed parameter-selection strategy. As described in
subsection 3.3, we random split the train samples
$D_m=(X_i,Y_i)_{i=1}^{500}$ into two disjoint equal size subsets, i.e., a learning set and a validation set. We first train on the learning
set $D^l_m$ to construct the $L_2$-RBoosting estimates
$f_{D^l_m,\alpha_k,k}$ and then use the validation set $D^v_m$ to
choose the appropriate shrinkage degree $\alpha^*_k$ and iteration
$k^*$ by minimizing the validation risk. Thirdly, we retrain the
obtained $\alpha^*_k$ on the entire training set $D_m$ to construct
$f_{D_m,\alpha^*_k,k}$ (Generally, if we have enough training
samples at hand, this step is optional). Finally, an independent
test set of 1000 noiseless observations are used to evaluate the
performance of $f_{D_m,\alpha^*_k,k}$.

Table \ref{t4}-Table \ref{t6} document the  test errors (RMSE) for
regression functions $m_1, \dots ,m_{9}$. The corresponding bold
numbers   denote the ideal generalization performance of the
$L_2$-RBoosting (choose optimal iteration $k^*$ and optimal
shrinkage degree $\alpha^*_k$ both according to minimize the test
error via the test sets). We also report the standard errors
(numbers in parentheses) of selected re-scale parameter $u$
over 20 independent runs in order to check the stability of such
parameter-selection strategy. From the tables, we can easily find
that the performance with such strategy approximates the
ideal one. More important, comparing the mean values and standard
errors of  $u$ with the  performance curves in
Fig.\ref{f1}-Fig.\ref{f3}, apart from  $m_8$,  we can distinctly
detect that the  selected  $u$  values by the proposed
parameter-selection strategy  are all located  near  the low valleys.

\begin{table}[H]
\begin{center}
 \caption{Performance comparison of $L_2$-Boosting, $L_2$-RBoosting and $L_2$-DDRBoosting on simulated regression data sets (1-dimension cases).}\label{t1}
\begin{tabular}{|c|c|c|c|}\hline
       & $m_1$  & $m_2$ & $m_3$     \\ \hline
  \multicolumn{4}{|c|}{$\sigma=0$} \\ \hline
  Boosting    &0.0318(0.0069)     &0.0895(0.0172)   &0.0184(0.0025)         \\ \hline
  RBoosting   &0.0308(0.0047)    & 0.0810(0.0183)   & 0.0179(0.0004)         \\ \hline
  DDRBoosting &\textbf{0.0268(0.0062)}    &\textbf{0.0747(0.0232)}   &\textbf{0.0178(0.0011)}         \\ \hline
 \multicolumn{4}{|c|}{$\sigma=0.5$} \\ \hline
  Boosting    &0.2203(0.0161)     &0.1925(0.0293)   &0.2507(0.0336)         \\ \hline
  RBoosting   & \textbf{0.2087(0.0181)}   & \textbf{0.1665(0.0210)}   & \textbf{0.2051(0.0252)}         \\ \hline
  DDRBoosting  &0.2388(0.0114)    &0.2142(0.0519)   &0.2508(0.0179)         \\ \hline
 \multicolumn{4}{|c|}{$\sigma=1$} \\ \hline
  Boosting    &0.3635(0.0467)      &0.2943(0.0375)   &0.3943(0.0415)         \\ \hline
  RBoosting   & \textbf{0.3479(0.0304)}     & \textbf{0.2558(0.0120)}   & \textbf{0.3243(0.0355)}         \\ \hline
  DDRBoosting  &0.3630(0.0315)    &0.3787(0.0614)   &0.4246(0.0360)         \\ \hline
 \end{tabular}
 \end{center}
 \end{table}

\begin{table}[H]
\begin{center}
 \caption{Performance comparison of $L_2$-Boosting, $L_2$-RBoosting and $L_2$-DDRBoosting on simulated regression data sets (2-dimension cases).} \label{t2}
\begin{tabular}{|c|c|c|c|}\hline
       & $m_4$   & $m_5$ & $m_6$   \\ \hline
  \multicolumn{4}{|c|}{$\sigma=0$} \\ \hline
  Boosting    &0.2125(0.0173)  & 0.2391(0.0140)   &0.3761(0.0235)            \\ \hline
  RBoosting   &\textbf{0.0582(0.0051)}  &\textbf{ 0.0930(0.0133)}   &\textbf{0.2161(0.0763)}           \\ \hline
  ARBoosting  &0.1298(0.0167)  & 0.1883(0.0216)   &0.3585(0.0573)          \\ \hline
 \multicolumn{4}{|c|}{$\sigma=0.5$} \\ \hline
  Boosting    &0.3646(0.0152)  & 0.3693(0.0111)   &0.4658(0.0233)           \\ \hline
  RBoosting   &\textbf{0.2392(0.0223)}  & \textbf{0.2665(0.0163)}   &\textbf{0.3738(0.0323)}            \\ \hline
  ARBoosting  &0.3439(0.0317)  & 0.3344(0.0117)   &0.5100(0.0788)            \\ \hline
 \multicolumn{4}{|c|}{$\sigma=1$} \\ \hline
  Boosting    &0.5250(0.0323)  & 0.3967(0.0317)   &0.5966(0.0424)            \\ \hline
  RBoosting   &\textbf{0.3836(0.0182)}  &\textbf{ 0.3759(0.0231)}   &\textbf{0.5066(0.0701)}           \\ \hline
  ARBoosting  &0.4918(0.0209)  & 0.4036(0.0180)   &0.5638(0.0450)          \\ \hline
 \end{tabular}
  \end{center}
 \end{table}

\begin{table}[H]
\begin{center}
 \caption{Performance comparison of $L_2$-Boosting, $L_2$-RBoosting and $L_2$-DDRBoosting on simulated regression data sets (10-dimension cases).}\label{t3}
\begin{tabular}{|c|c|c|c|}\hline
       & $m_7$   & $m_8$ & $m_{9}$      \\ \hline
  \multicolumn{4}{|c|}{$\sigma=0$} \\ \hline
  Boosting    &1.4310(0.0412)    &0.4167(0.0324)   &0.8274(0.0142)         \\ \hline
  RBoosting   &\textbf{0.7616(0.0144)}   &\textbf{0.4167(0.0403)}   &\textbf{0.6875(0.0107)}         \\ \hline
  DDRBoosting  &1.1322(0.0696)    &0.4178(0.0409)   &0.8130(0.0330)         \\ \hline
 \multicolumn{4}{|c|}{$\sigma=0.5$} \\ \hline
  Boosting    &1.4450(0.0435)   &0.4283(0.0314)   &0.8579(0.0414)         \\ \hline
  RBoosting   &\textbf{0.7755(0.0475)}    &\textbf{0.4283(0.0401) }   &\textbf{0.7218(0.0223)}         \\ \hline
  DDRBoosting  &1.2526(0.0290)     & 4381(0.0304) &0.8385(0.0222)         \\ \hline
 \multicolumn{4}{|c|}{$\sigma=1$} \\ \hline
  Boosting    &1.4420(0.0413)     &0.4404(0.0242)   &0.8579(0.0415)         \\ \hline
  RBoosting   &\textbf{0.8821(0.0575)}    &\textbf{0.4404(0.0321)}   &\textbf{0.8406(0.0175)}         \\ \hline
  DDRBoosting  &1.4423(0.0625)    &0.4503(0.0393)   &0.9295(0.0120)         \\ \hline
 \end{tabular}
 \end{center}
 \end{table}

\begin{table}[H]
\addtolength{\tabcolsep}{-6pt}
\begin{center}
 \caption{Performance of $L_2$-RBoosting via parameter-selection strategy on simulated regression data sets (1-dimension case).}\label{t4}
\begin{tabular}{|c|c|c|c|c|c|c|}\hline
                 & $m_1$  & $u$ & $m_2$ & $u$ & $m_3$ & $u$     \\ \hline
  \multirow{2}*{$\sigma=0$}     &0.0317(0.0069) & 158(164)     &0.0791(0.0262) & $5(2)$   &0.0180(0.0013) & $232(95)$         \\
                     &\textbf{0.0308(0.0062)} &        &\textbf{0.0747(0.0232)} & $ $     &\textbf{0.0178(0.0011)} & $ $         \\ \hline
  \multirow{2}*{$\sigma=0.5$}   &0.2113(0.0122) & 609(160)         &0.1766(0.0135) & $3(3)$   &0.2118(0.0094) & $3(2)$           \\
                                &\textbf{0.2087(0.0181)} &     &\textbf{0.1665(0.0210)} & $  $    &\textbf{0.2051(0.0252)} & $ $   \\ \hline
  \multirow{2}*{$\sigma=1$}     &0.3487(0.0132) & $987(440)$          &0.2800(0.0308) & $1(0)  $ &0.3302(0.0511) & $148(400)$         \\
                               &\textbf{0.3479(0.0304)}&  &\textbf{0.2558(0.0120)} & $   $   &\textbf{0.3243(0.0355)} & $$           \\ \hline
\end{tabular}
\end{center}
\end{table}

\begin{table}[H]
\addtolength{\tabcolsep}{-6pt}
\begin{center}
 \caption{Performance of $L_2$-RBoosting via parameter-selection strategy on simulated regression data sets (2-dimension case).} \label{t5}
\begin{tabular}{|c|c|c|c|c|c|c|}\hline
                 & $m_4$  & $u$ & $m_5$ & $u$ & $m_6$ & $u$      \\ \hline
  \multirow{2}*{$\sigma=0$}     &0.0593(0.0059) & 55(34)    & 0.0958(0.0063) & $10(8)$     &0.2210(0.0143) & $3(2)$           \\
                                &\textbf{0.0582(0.0051)} &       &\textbf{0.0930(0.0133)} & $ $ &\textbf{0.2161(0.0763)} & $$           \\ \hline
  \multirow{2}*{$\sigma=0.5$}   &0.2511(0.0130) & 4(3)     & 0.2848(0.0201) & $142(300)$    &0.3869(0.0173) & $20(30)$           \\
                                 &\textbf{0.2392(0.0223)} &     & \textbf{0.2665(0.0163)} & $$     &\textbf{0.3738(0.0323)} & $$           \\ \hline
 \multirow{2}*{$\sigma=1$}     &0.4001(0.0179) & $6(7)$   & 0.4007(0.0170) & $2(1)$      &0.5123(0.0925) & $6(7)$           \\
                             &\textbf{0.3836(0.0182)} & &\textbf{0.3759(0.0231)} & $ $   &\textbf{0.5066(0.0701)} & $$           \\ \hline
 \end{tabular}
  \end{center}
 \end{table}

\begin{table}[H]
\addtolength{\tabcolsep}{-4.5pt}
\begin{center}
 \caption{Performance of $L_2$-RBoosting via parameter-selection strategy on simulated regression data sets (10-dimension case).} \label{t6}
\begin{tabular}{|c|c|c|c|c|c|c|}\hline
                 & $m_7$  & $u$ & $m_{8}$ & $u$ & $m_{9}$ & $u$     \\ \hline
  \multirow{2}*{$\sigma=0$}     &0.7765(0.0259)         & 66(65)&0.4169(0.0313)          & $\diagdown$   &0.6882(0.0113)          & $42(36)$         \\
                                &\textbf{0.7616(0.0144)}&       &\textbf{0.4167(0.0403)} &            &\textbf{0.6875(0.0107)} &          \\ \hline
  \multirow{2}*{$\sigma=0.5$}   &0.7757(0.0128)         & 72(59)&0.4349(0.0335)          & $\diagdown$   &0.7396(0.0145)          & $11(9)$           \\
                                &\textbf{0.7755(0.0475)}&       &\textbf{0.4283(0.0401)} & $ $ &\textbf{0.7218(0.0223)} & $$         \\ \hline
  \multirow{2}*{$\sigma=1$}     &0.9093(0.0304)  & $38(37)$&0.4452(0.0308) & $\diagdown$   &0.8539(0.0278) & $23(30)$         \\
                                &\textbf{0.8821(0.0575)}  & &\textbf{0.4404(0.0321)} & $ $&\textbf{0.8406(0.0175)} & $$         \\ \hline

 \end{tabular}
  \end{center}
 \end{table}

\subsection{Real data experiments}

We have verified that  $L_2$-RBoosting  outperforms $L_2$-Boosting
and $L_2$-DDRBoosting on the $3\times 9=27$ different distributions in the
previous simulations. We now further compare the learning
performances of these boosted-type algorithms on six real data sets.

The  first data set is a subset of the Shanghai Stock Price Index (SSPI),
which can be extracted from http://www.gw.com.cn. This data set
contains 2000 trading days' stock index which records five
independent variables, i.e., maximum price, minimum price, closing
price,  day trading quota,  day trading volume, and one dependent
variable, i.e., opening price.  The second one is the Diabetes
data set\cite{Efron2004}. This data set contains 442 diabetes
patients that were measured on ten independent variables, i.e., age, sex,
body mass index etc. and one response variable, i.e., a measure of disease
progression. The third one is  the Prostate cancer data set derived  from
a study of prostate cancer by Blake et al.\cite{Blake1998}. The data
set consists of the medical records of 97 patients who were about to
receive a radical prostatectomy. The predictors are eight clinical
measures, i.e., cancer volume, prostate weight, age etc. and one response variable, i.e., the logarithm of
prostate-specific antigen. The fourth one is the Boston Housing data set
created form a housing values survey in suburbs of Boston by
Harrison\cite{Harrison1978}. This data set contains 506 instances
which include thirteen attributions, i.e., per capita crime rate by
town, proportion of non-retail business acres per town, average
number of rooms per dwelling etc. and one response variable, i.e.,
median value of owner-occupied homes. The fifth one is the Concrete
Compressive Strength (CCS) data set created
from\cite{Ye1998}.  The data set contains 1030 instances including
eight quantitative independent variables, i.e., age and ingredients etc.
and one dependent variable, i.e., quantitative concrete compressive
strength. The sixth one is the Abalone data set, which comes from an original
study in \cite{Nash1994} for predicting the age of abalone from
physical measurements. The data set  contains 4177  instances which
were measured on eight independent variables, i.e., length, sex,
height etc. and one response variable, i.e., the number of
rings.

Similarly, we divide  all the real data sets  into
two disjoint equal parts (except for the Prostate Cancer
data set, which were divided into two parts beforehand: a training
set with 67 observations and a test set with 30 observations). The first half serves as the training set
and the second half serves as the test set. For each real data experiment, weak learners are changed  to the  decision stumps
(specifying one split of each tree, $J=1$) corresponding to an additive model with only main effects. Table \ref{t7} documents
the performance (test RMSE) comparison results of $L_2$-Boosting,
$L_2$-RBoosting and $L_2$-DDRBoosting on six real data sets,
respectively (the bold numbers denote the optimal performance). It is observed from the table that the performance of
$L_2$-RBoosting with $u$ selected via our recommended strategy
outperforms both $L_2$-Boosting and $L_2$-DDRBoosting on
all  real data sets, especially for some data sets, i.e., Diabetes, Prostate and CCS, makes a great improvement.



\begin{table}[H]
\addtolength{\tabcolsep}{-4pt}
\begin{center}
 \caption{Performance comparison of $L_2$-Boosting, $L_2$-RBoosting and $L_2$-DDRBoosting
on  real data sets}\label{para} \label{t7}
 \begin{tabular}{|c|c|c|c|c|c|c|}\hline
 \backslashbox[2cm] {Methods}{Datasets} & Stock  & Diabetes & Prostate & Housing & CCS  & Abalone                            \\ \hline
 Boosting    &0.0050          &60.5732          &0.6344           & 0.6094           & 0.7177            & 2.1635                   \\ \hline
 RBoosting   &\textbf{0.0047} &\textbf{55.0137} &\textbf{0.4842}  & \textbf{0.6015}  & \textbf{0.6379}   &  \textbf{2.1376}         \\ \hline
 DDRBoosting &0.0049          &59.3595          &0.6133           & 0.6281           & 0.6977            &   2.1849       \\ \hline
  \end{tabular}
 \end{center}
 \end{table}

\section{Proofs}

In this section, we provide the proofs of the main results. At
first, we aim to prove Theorem \ref{THEOREM1}. To this end, we
shall give an error decomposition strategy for $\mathcal
E(\pi_Mf_{k})-\mathcal E(f_\rho)$. Using the similar methods that
in \cite{Lin2013,Xu2015a}, we construct an  $f_k^*\in
\mbox{span}(D_n)$ as follows. Since $f_\rho\in \mathcal L_1^r$,
there exists a $h_\rho:=\sum_{i=1}^na_ig_i\in \mbox{span}(S)$ such
that
\begin{equation}\label{h}
                \|h_\rho\|_{\mathcal L_1}\leq\mathcal B,\
                \mbox{and}\ \|f_\rho-h_\rho\|\leq \mathcal B n^{-r}.
\end{equation}
Define
\begin{equation}\label{f*}
               f_0^*=0,\
               f_k^*=\left(1-\frac1k\right)f^*_{k-1}+\frac{\sum_{i=1}^n|a_i|\|g_i\|_\rho}{k}g^*_k,
\end{equation}
 where
$$
              g_k^*:=\arg\max\limits_{g\in S'}\left\langle
            h_\rho-\left(1-\frac1k\right)f_{k-1}^*,g\right\rangle_{\rho},
$$
and
$$
            \mathcal D_n':=\left\{{g_i(x)}/{\|g_i\|_\rho}\right\}_{i=1}^n
             \bigcup
                 \left\{-{g_i(x)}/{\|g_i\|_\rho}\right\}_{i=1}^n
$$
with $g_i\in S$.

 Let $f_{k}$  and $f_k^*$ be defined as in Algorithm \ref{alg2} and   (\ref{f*}), respectively. We
have
\begin{eqnarray*}
         &&\mathcal E(\pi_Mf_{ k})-\mathcal E(f_\rho)\\
         &\leq&
         \mathcal E(f_k^*)-\mathcal E(f_\rho)
         +
         \mathcal E_{\bf z}(\pi_Mf_{ ,k})-\mathcal E_{D}(f_k^*)\\
         &+&
         \mathcal
        E_{D}(f_k^*)-\mathcal E(f_k^*)+\mathcal E(\pi_Mf_{ k})-\mathcal
        E_D(f_{ k}).
\end{eqnarray*}
Upon making the short hand notations
$$
             \mathcal D(k):=\mathcal E(f_k^*)-\mathcal E(f_\rho),
$$
 $$
          \mathcal S(D,k):=\mathcal
           E_{D}(f_k^*)-\mathcal E(f_k^*)+\mathcal E(\pi_Mf_{ k})-\mathcal
           E_{D}(\pi_Mf_{ k}),
$$
and
$$
            \mathcal P(D,k):=\mathcal E_{D}(\pi_Mf_{ k})
            -\mathcal E_{D}(f_k^*)
$$
respectively for the approximation error,  sample error and
hypothesis error, we have
\begin{equation}\label{error decomposition}
            \mathcal E(\pi_Mf_{ k})-\mathcal E(f_\rho)=\mathcal
            D(k)+ \mathcal S(D,k)+\mathcal P(D,k).
\end{equation}

To bound estimate   $\mathcal D(k)$, we need the following Lemma
\ref{LEMMA1}, which can be found in \cite[Prop.1]{Lin2013}.

\begin{lemma}\label{LEMMA1}
 Let $f_k^*$ be defined in (\ref{f*}). If
$f_\rho\in \mathcal L_1^r$, then
\begin{equation}\label{approximation error estimation}
             \mathcal D(k)\leq  \mathcal B^2(k^{-1/2}+n^{-r})^2.
\end{equation}
\end{lemma}

 To bound the hypothesis error, we need the following  two lemmas. The first one can be
found in \cite{Lin2015}, which is a direct
 generalization of \cite[Lemma 2.3]{Temlyakov2008}.

\begin{lemma}\label{NUMBER THEORY}
Let $j_0>2$ be a natural number. Suppose that three positive numbers
$c_1<c_2\leq j_0$, $\mathcal C_0$ be given. Assume that a sequence
$\{a_n\}_{n=1}^\infty$ has the following two properties:

(i) For all $1\leq n\leq j_0$,
$$     a_n\leq \mathcal C_0 n^{-c_1},
$$
 and,
for all $n\geq j_0$,
$$
        a_n\leq a_{n-1}+\mathcal C_0(n-1)^{-c_1}.
$$

(ii) If for some $v\geq j_0$ we have
$$
             a_v\geq \mathcal C_0v^{-c_1},
$$
then
$$
            a_{v+1}\leq a_v(1-c_2/v).
$$

Then, for all $n=1,2,\dots,$ we have
$$
            a_n\leq   2^{1+\frac{c_1^2+c_1}{c_2-c_1}} \mathcal
            C_0n^{-c_1}.
$$
\end{lemma}

The second one can be easily deduced from \cite[Lemma
2.2]{Temlyakov2008}.

\begin{lemma}\label{IMPORTANT ESTIMATE}
Let $h\in\mbox{span}(S),$   $f_{,k}$ be the estimate defined in
Algorithm \ref{alg2} and $y(\cdot)$ is an arbitrary function satisfying
$y(x_i)=y_i$. Then, for arbitrary $k=1,2,\dots$, we have
\begin{equation*}
\begin{aligned}
      & \|f_{ k}-y\|_m
      \leq
      \|f_{ k-1}-y\|_m \\
     & \left( \! 1\! - \! \alpha_k \! \left(1\! - \! \frac{\|y-h\|_m}{\|f_{  k-1}-y\|_m} \! \right)
      \! + \! 2\left(\frac{\alpha_k(\|y\|_m+ \!
      \|h\|_{\mathcal L_1(S)})}{(1-\alpha_k)\|f_{ k-1}-y\|_m}\right)^2 \! \right).
\end{aligned}
\end{equation*}
\end{lemma}

Now, we are in a position to present the hypothesis error estimate.

\begin{lemma}\label{HYPOTHESIS ERROR}
Let $f_{ k}$ be the estimate defined in Algorithm \ref{alg2}. Then, for
arbitrary $h\in\mbox{span}(S)$ and $u\in\mathcal N$, there holds
$$
         \|f_{ k}-y\|_m^2\leq 2\|y-h\|_m^2+2(M+\|h\|_{\mathcal L_1(S)})^2
         2^\frac{3u^2+14u+20}{8u+8}k^{-1}.
$$
\end{lemma}

\begin{IEEEproof}
 By Lemma \ref{IMPORTANT ESTIMATE}, for $k\geq 1$, we obtain
\begin{equation*}
\begin{aligned}
     & \|f_{k}-y\|_m-\|y-h\|_m  \leq \\ & (1-\alpha_k)(\|f_{ k-1}-y\|_m-\|y-h\|_m)
        \\ & + C\|f_{ k-1}-y\|_m\left(\frac{\alpha_k(\|y\|_m
      +\|h\|_{\mathcal L_1(S)})}{\|f_{ k-1}-y\|_m}\right)^2.
\end{aligned}
\end{equation*}
Let
$$
      a_{k+1}=\|f_{k}-y\|_m-\|y-h\|_m.
$$
Then, by noting $\|y\|_m\leq M$, we have
$$
      a_{k+1}\leq (1-\alpha_k)a_{k}+C \frac{\alpha_k^2(M+\|h\|_{\mathcal L_1(S)})^2}{a_{k}}.
$$
We plan to apply Lemma \ref{NUMBER THEORY} to the sequence
$\{a_n\}$. Let $\mathcal
C_0=\max\{1,\sqrt{C}\}2(M+\|h\|_{l^1(\mathcal
      D_n)})$
According to the definitions of $\{a_k\}_{k=1}^\infty$ and $f_{ k}$,
we obtain
$$
            a_1=\|y\|_m-\|y-h\|_m\leq 2M+\|h\|_{\mathcal L_1(S)}\leq \mathcal C_0,
$$
and
$$
           a_{k+1}\leq a_k+\alpha_k\|y\|_m\leq a_k+\mathcal
           C_0k^{-1/2}.
$$
Let   $a_k\geq \mathcal C_0k^{-1/2}$, since $\alpha_k=\frac2{k+u}$,
we then obtain
$$
          a_k\leq
          \frac{k+u-2}{k+u}a_{k-1}+Ca_{k-1}k\frac4{\mathcal C_0^2(k+u)^2}
          (M+\|h\|_{\mathcal L_1(S)})^2.
$$
That is,
\begin{equation*}
\begin{aligned}
      &  a_k\leq a_{k-1}\left(1-\frac{2}{k+u}+C\frac{4k}{\mathcal C_0^2(k+u)^2}
        (M+\|h\|_{\mathcal L_1(S)})^2\right)
       \\ & \leq  (1-\left(\frac12+\frac{2u+2}{(2+u)^2}\right)\frac1{k-1}).
\end{aligned}
\end{equation*}
Now, it follows from Lemma \ref{NUMBER THEORY} with $c_1=\frac12$
and $c_2=\frac12+\frac{2u+2}{(2+u)^2}$ that
$$
            a_n\leq  \max\{1,\sqrt{C}\}2(M+\|h\|_{\mathcal L_1(S)})
            2^{1+\frac{3(u+2)^2}{8u+8}}
             n^{-1/2}.
$$
Therefore, we obtain
$$
         \|f_{k}-y\|_m\leq \|y-h\|_m+(M+\|h\|_{\mathcal L_1(S)})
         2^\frac{3u^2+14u+20}{8u+8}k^{-1/2}.
$$
This finishes the proof of Lemma \ref{HYPOTHESIS ERROR}.
\end{IEEEproof}


Now we proceed the proof of Theorem \ref{THEOREM1}.

\begin{IEEEproof}[Proof of Theorem \ref{THEOREM1}]
 Based on Lemma \ref{HYPOTHESIS ERROR} and the fact
$\|f^*_k\|_{\mathcal L_1(S)}\leq \mathcal B$ \cite[Lemma
1]{Lin2013}, we obtain
\begin{equation}\label{hypothesis error estimation}
            \mathcal P(D,k)\leq 2
            \mathcal E_{D}(\pi_Mf_{ k})-
            \mathcal E_{D}(f_k^*)\leq 2(M+\mathcal B)^22^\frac{3u^2+14u+20}{8u+8}k^{-1}.
\end{equation}

Therefore, both the approximation error and hypothesis error are
deduced. The only thing remainder is to bound
 bound the sample error $\mathcal S(D,k)$.
Upon using the short hand notations
$$
               S_1(D,k):=\{\mathcal E_{D}(f_k^*)-\mathcal E_{D}(f_\rho)\}
               -\{\mathcal E(f_k^*)-\mathcal
               E(f_\rho)\}
$$
and
$$
               S_2(D,k):=\{\mathcal E(\pi_Mf_{ k})-\mathcal E(f_\rho)\}
               -\{\mathcal E_{D}(\pi_Mf_{ k})-\mathcal E_{D}(f_\rho)\},
$$
we write
\begin{equation}\label{sample decomposition}
            \mathcal S(D,k)=\mathcal S_1(D,k)+\mathcal
            S_2(D,k).
\end{equation}
It can be found in   \cite[Prop.2]{Lin2013} that
 for any $0<t<1$, with confidence
$1-\frac{t}2$,
\begin{equation}\label{S1 estimate}
              \mathcal S_1(D,k)\leq \frac{7(3M+\mathcal B\log\frac2t)}{3m}+\frac12\mathcal D(k)
\end{equation}
It also follows from  \cite[Eqs(A.10)]{Xu2014} that
\begin{equation}\label{S2 estimate}
      \mathcal S_2(D,k)\leq \frac12\mathcal E(\pi_Mf_{ k})-\mathcal
      E(f_\rho)+\log\frac2t\frac{Ck\log
      m}{m}
\end{equation}
holds with confidence at least $1-t/2$. Therefore, (\ref{error
decomposition}), (\ref{approximation error estimation}),
(\ref{hypothesis error estimation}), (\ref{sample decomposition}),
(\ref{S1 estimate}) and (\ref{S2 estimate})   yield  that
\begin{equation*}
\begin{aligned}
 & \mathcal E(\pi_Mf_{ k})-\mathcal E(f_\rho)
             \leq
            \\ &  C(M\! +\! \mathcal B)^2 \left(2^\frac{3u^2+14u+20}{8u+8}k^{-1}
            \! +   \! (m/k)^{-1}\log m\log\frac2t+n^{-2r} \! \right)
\end{aligned}
\end{equation*}
holds with confidence at least $1-t$. This finishes the proof of
Theorem \ref{THEOREM1}.
\end{IEEEproof}

\begin{IEEEproof}[Proof of Theorem \ref{THEOREM2}] It can be
  deduced from \cite[Theorem 1.2]{Temlyakov2008} and the
same method as in the proof of Theorem \ref{THEOREM1}. For the sake
of brevity, we omit the details.
\end{IEEEproof}

\begin{IEEEproof}[Proof of Proposition \ref{Proposition1}] It is easy to check that
$$
         f_{ k}=(1-\alpha_k) f_{ k-1}+\langle
        y-(1-\alpha_k) f_{ k-1},g_k\rangle_2g_k.
$$
As $\|g_k\|\leq 1$, we obtain from the H\"{o}lder inequality that
\begin{equation*}
\begin{aligned}
   &   \langle
        y-(1-\alpha_k) f_{ k-1},g_k\rangle_2\leq \|y-(1-\alpha_k) f_{ k-1}\|_2
        \\ & \leq
        (1-\alpha_k)\|y-f_{ k-1}\|_2+\alpha_kM.
\end{aligned}
\end{equation*}
As
$$
         \|y-f_{ k-1}\|_2\leq C(M+\|h\|_{\mathcal L_1(S)})k^{-1/2}+n^{-r},
$$
we can obtain
$$
        \|f_{ k}\|_1\leq C((M+\|h\|_{\mathcal L_1(S)})k^{1/2}+kn^{-r}).
$$
This finishes the proof of Proposition \ref{Proposition1}.
\end{IEEEproof}

 Now we turn to prove Theorem \ref{THEOREM3}. The following
 concentration inequality \cite{Wu2007} plays a crucial role in the
 proof.

\begin{lemma}\label{CONCERTRATION INEQUA}
  Let $\mathcal F$ be a
class of measurable functions on $Z$. Assume that there are
constants $B,c>0$ and $\alpha\in[0,1]$ such that $\|f\|_\infty\leq
B$ and $\mathbf Ef^2\leq c(\mathbf E f)^\alpha$ for every
$f\in\mathcal F.$ If for some $a>0$ and $\mu\in(0,2)$,
\begin{equation}\label{condition}
                  \log\mathcal N_2(\mathcal F,\varepsilon)\leq
                  a\varepsilon^{-\mu},\ \ \forall\varepsilon>0,
\end{equation}
then there exists a constant $c_p'$ depending only on $p$ such that
for any $t>0$, with probability at least $1-e^{-t}$, there holds
\begin{eqnarray}\label{lemma3}
                 & \mathbf Ef-\frac1m\sum_{i=1}^mf(z_i)\leq  \\
                 & \frac12\eta^{1-\alpha}(\mathbf
                 Ef)^\alpha+c_\mu'\eta\nonumber+
                 2\left(\frac{ct}{m}\right)^\frac1{2-\alpha}
                 +
                 \frac{18Bt}{m},\
                 \forall f\in\mathcal F,
\end{eqnarray}
where
$$
               \eta:=\max\left\{c^\frac{2-\mu}{4-2\alpha+\mu\alpha}\left(\frac{a}m\right)^\frac2{4-2\alpha+\mu\alpha},
               B^\frac{2-\mu}{2+\mu}\left(\frac{a}m\right)^\frac2{2+\mu}\right\}.
$$
\end{lemma}

We continue the proof of Theorem \ref{THEOREM3}.

\begin{IEEEproof}[Proof of Theorem \ref{THEOREM3}]
 For arbitrary $h\in\mbox{span}(S)$,
\begin{equation*}
\begin{aligned}
             \mathcal E(f_{k})-\mathcal E(h)
             = & \mathcal E(f_{ k})-\mathcal E(h)-
             (\mathcal E_{D}(f_{ k})-\mathcal E_{D}(h))
            \\ & +\mathcal E_{D}(f_{ k})-\mathcal E_{D}(h).
\end{aligned}
\end{equation*}
  Set
\begin{equation}\label{space}
        \mathcal G_{R}:=\left\{g(z)=(\pi_Mf(x)-y)^2- (h(x)-y)^2:f\in
        B_R\right\}.
\end{equation}
 Using the obvious inequalities $\|\pi_Mf\|_\infty \leq M$,  $|y|\leq
M$ a.e.,  we get the inequalities
$$
            |g(z)|\leq (3M+\|h\|_{\mathcal L_1(S)})^2
$$
and
$$
           \mathbf
           Eg^2
           \leq
           (3M+\|h\|_{\mathcal L_1(S)})^2\mathbf Eg.
$$
For $g_1,g_2\in\mathcal G_R$, it follows   that
$$
          |g_1(z)-g_2(z)|
           \leq
           (3M+\|h\|_{\mathcal L_1(S)})|f_1(x)-f_2(x)|.
$$
Then
\begin{equation*}
\begin{aligned}
    &     \mathcal N_{2}(\mathcal G_R,\varepsilon)
         \leq
          \mathcal N_{2,{\bf
         x}}\left(
         B_R,\frac\varepsilon{3M+\|h\|_{\mathcal L_1(S)}
         }\right)
        \\ & \leq
         \mathcal N_{2,{\bf
         x}}\left(
         B_1,\frac\varepsilon{ R(3M+\|h\|_{\mathcal L_1(S)})
         }\right).
\end{aligned}
\end{equation*}
Using the above inequality and Assumption \ref{Assumption1}, we have
$$
            \log \mathcal N_{2}(\mathcal F_R,\varepsilon)\leq
            \mathcal L( R(3M+\|h\|_{\mathcal L_1(S)}))^\mu\varepsilon^{-\mu}.
$$
By Lemma \ref{CONCERTRATION INEQUA} with $B=c= (3M+\|h\|_{\mathcal
L_1(S)})^2$, $\alpha=1$ and $a=\mathcal L( R(3M+\|h\|_1))^\mu$, we
know that for any $t\in (0,1),$ with confidence $1-\frac{t}2,$ there
exists a constant $C$ depending only on $d$  such that for all
$g\in\mathcal G_R$
$$
        \mathbf Eg-\frac1m\sum_{i=1}^mg(z_i)
        \leq
        \frac12\mathbf
        Eg+c_\mu'\eta+20(3M+\|h\|_{\mathcal L_1(S)})^2\frac{\log4/t}{m}.
$$
Here
$$
             \eta=((3M+\|h\|_{\mathcal L_1(S)})^2)^{\frac{2-\mu}{2+\mu}}
             \left(\frac{\mathcal L( R(3M+\|h\|_{\mathcal L_1(S)}))^\mu}{m}
             \right)^{\frac{2-\mu}{2+\mu}}.
$$

It then follows from Proposition \ref{Proposition1} that
\begin{equation*}
\begin{aligned}
          &   \mathcal
              E(f_{  k})-\mathcal E(f_\rho)
           \leq \\ &
         C\log\frac2t(3M+\mathcal
         B)^2   \left(n^{-2r}+k^{-1}+\left(\frac{(kn^{-r}+\sqrt{k})^\mu}{m}\right)^{\frac{2-\mu}{2+\mu}}
         \right).
\end{aligned}
\end{equation*}
This finishes the proof of Theorem \ref{THEOREM3}.
\end{IEEEproof}

\section{Conclusion}


In this paper, we draw a concrete analysis concerning how to
determine the shrinkage degree into $L_2$-RBoosting. The
contributions can be concluded in four aspects.
 Firstly, we
theoretically deduced the generalization error bound of
$L_2$-RBoosting and demonstrated the importance of the shrinkage
degree. It is shown that, under certain conditions, the learning rate
of $L_2$-RBoosting can reach $O(m^{-1/2}\log m)$, which is the same
as the optimal ``record'' for greedy learning and boosting-type
algorithms. Furthermore, our result showed that although the
shrinkage degree did not affect the learning rate, it determined the
constant of the generalization error bound, and therefore, played a
crucial role in $L_2$-RBoosting learning with finite samples. Then, we proposed two schemes to determine the
shrinkage degree. The first one is the conventional parameterized
$L_2$-RBoosting,  and the other one is to learn the shrinkage degree
 from the  samples directly ($L_2$-DDRBoosting). We further provided the theoretically
optimality of these approaches. Thirdly,
 we compared these two
approaches and proved that, although $L_2$-DDRBoosting reduced the
parameters, the estimate deduced from $L_2$-RBoosting possessed a
better structure ($l^1$ norm). Therefore, for some special weak
learners, $L_2$-RBoosting can achieve better performance than
$L_2$-DDRBoosting. Fourthly,  we   developed an adaptive
parameter-selection strategy for the shrinkage degree. Our
theoretical results demonstrated that, $L_2$-RBoosting with such a
shrinkage degree selection strategy did not degrade the
generalization capability very much. Finally, a series of numerical
simulations and real data experiments have been carried out to verify our
 theoretical assertions. The obtained results enhanced the
 understanding of RBoosting
and could provide guidance on how to utilize $L_2$-RBoosting for regression tasks.



%

%
%
%
%
%

\ifCLASSOPTIONcaptionsoff
  \newpage
\fi

\bibliographystyle{IEEEtran}
\bibliography{IEEEabrv,KylinRef}

\end{document}